\RequirePackage{fix-cm}
\documentclass[smallextended]{svjour3}       
\smartqed  
\usepackage{graphicx}
\usepackage{natbib}
\usepackage{amsfonts}
\usepackage{amsmath}
\usepackage{hyperref}
\usepackage{mathptmx}
\usepackage{latexsym}
\usepackage{siunitx}
\usepackage{booktabs}
\usepackage{caption}
\usepackage{multirow}
\usepackage{multicol}
\usepackage[svgnames]{xcolor}

\usepackage{color}

\definecolor{yellow_annot}{RGB}{187,187,1}
\definecolor{green_annot}{RGB}{141,179,74}
\definecolor{orange_annot}{RGB}{210,140,65}
\definecolor{blue_annot}{RGB}{104,170,158}

\newcommand{\co}[1]{\textcolor{yellow_annot}{#1}}
\newcommand{\ca}[1]{\textcolor{green_annot}{#1}}
\newcommand{\e}[1]{\textcolor{orange_annot}{#1}}
\newcommand{\rs}[1]{\textcolor{blue_annot}{#1}}

\usepackage{todonotes}

\journalname{Language Resources \& Evaluation}

\begin{document}
\thanks{This research received funding from the Flemish Government under the Research Program Artificial Intelligence - 174L00121 (2021). The research also received funding from the Research Foundation Flanders (FWO-Vlaanderen) with grant number 1S96322N.}

\title{EmoTwiCS: A Corpus for Modelling Emotion Trajectories in Dutch Customer Service Dialogues on Twitter}

\titlerunning{EmoTwiCS: A Corpus for Modelling Emotion Trajectories}

\authorrunning{Sofie Labat et al.}

\author{Sofie Labat$^{\diamondsuit}$ \and Thomas Demeester$^{\clubsuit}$ \and Véronique Hoste$^{\diamondsuit}$}

\institute{$^{\diamondsuit}$ 
            LT3 Language and Translation Technology Team, Department of Translation, Interpreting and Communication, Ghent University, Groot-Brittaniëlaan 45, 9000 Ghent, Belgium.\\
            $^{\clubsuit}$ 
            T2K Group, IDLab, Department of Information Technology, Ghent University - Imec Belgium, Technologiepark 126, 9052 Ghent, Belgium.\\ \\
            \email{\{Sofie.Labat, Veronique.Hoste, Thomas.Demeester\}@UGent.be}}


\maketitle
\begin{abstract}

Due to the rise of user-generated content, social media is increasingly adopted as a channel to deliver customer service. Given the public character of these online platforms, the automatic detection of emotions forms an important application in monitoring customer satisfaction and preventing negative word-of-mouth. This paper introduces EmoTwiCS, a corpus of 9,489 Dutch customer service dialogues on Twitter that are annotated for emotion trajectories. In our business-oriented corpus, we view emotions as dynamic attributes of the customer that can change at each utterance of the conversation. The term `emotion trajectory' refers therefore not only to the fine-grained emotions experienced by customers (annotated with 28 labels and valence-arousal-dominance scores), but also to the event happening prior to the conversation and the responses made by the human operator (both annotated with 8 categories). Inter-annotator agreement (IAA) scores on the resulting dataset are substantial and comparable with related research, underscoring its high quality. Given the interplay between the different layers of annotated information, we perform several in-depth analyses to investigate (i) static emotions in isolated tweets, (ii) dynamic emotions and their shifts in trajectory, and (iii) the role of causes and response strategies in emotion trajectories. We conclude by listing the advantages and limitations of our dataset, after which we give some suggestions on the different types of predictive modelling tasks and open research questions to which EmoTwiCS can be applied. The dataset is available upon request and will be made publicly available upon acceptance of the paper. 


\end{abstract}

\section{Introduction}\label{sec1}

The rapid adoption of second generation web-based applications and the corresponding rise of user-generated content are causing radical changes to customer behaviours. Contemporary customers' time is scarce; they are self-educated through social media; they expect a personalized service from authentic and sustainable companies; and they are more likely to churn if their expectations are not satisfied~\citep{DeloitteDig19}. As a result, organizations have to rethink their traditional business models to meet these novel customer needs~\citep{Hennig-Thurau2010}, thus causing a paradigm shift in the field of customer relationship management (CRM)~\citep{Greenberg2009}. 
Customer service is a crucial tool for CRM and long-lasting business success: it increases customer satisfaction, which has a positive effect on customer loyalty and customer retention~\citep{Dresner95,Gustafsson2005,Hossain2013}. As customer preferences evolve, the provision of customer service on public channels, such as social media platforms, is gaining ground. This trend is beneficial to customers who appreciate the transparency and immediacy of online media, but also to businesses, as social media-based customer service results in higher reputation scores~\citep{Guo2020}.

The evolution gives, however, also rise to novel challenges, since publicly shared complaints are more difficult to control from a company's perspective~\citep{Gallaugher2010}, considering the wide and rapid dissemination of content on social media. In an effort to monitor their customers on social media and provide timely responses, more and more firms are investing in webcare teams~\citep{vanNoort2012}, the automatic analysis of textual data~\citep{Berger2020}, and the development of conversational agents~\citep{Ngai21}. In this respect, the automatic analysis of customer emotions during social media interactions forms an important application, as it can be applied to identify customers who urgently need help, track customer satisfaction, reduce churns (i.e., customers leaving the company), and monitor the (un)successfulness of a customer service interaction. Moreover, knowledge about customer emotions and possible response strategies can also be implemented in conversational agents operating on social media channels to create emotion-aware assistants.

Nevertheless, most studies on emotion recognition in conversations (ERC) rely on open-domain conversations from publicly released datasets that are `artificial' in the sense that they contain mock-up conversations extracted from, e.g., English learner websites~\citep{Li2017} or TV show subtitles~\citep{Chen2018}. Unfortunately, such datasets are not tailored towards specific industrial applications~\citep{Guibon2021} that generally encounter more noisy, imbalanced and domain-specific data. Our research aims to address this shortcoming by introducing EmoTwiCS, a novel natural language processing (NLP) resource designed for the task of modelling emotion trajectories in Dutch customer service dialogues on Twitter. In contrast to existing datasets in the field of ERC, we propose to model fine-grained emotion trajectories in a closed-domain, business-related conversational setting. The term \textit{fine-grained} refers to the way in which emotions are annotated along a large categorical taxonomy and with dimensional \textit{valence}-\textit{arousal}-\textit{dominance} scores. The term \textit{trajectories} hints at the fact that we regard emotions as dynamic attributes of the customer that can change at each utterance of the conversation. To better understand how such changes occur, we also consider (i) the event happening prior to the conversation and (ii) the response strategies applied by customer service agents as part of the trajectory. Finally, as there currently exists only one small artificial dataset of 11 conversations for the task of ERC in the Dutch language~\citep{Vaassen2012}, the EmoTwiCS dataset fills this research gap by introducing a much larger dataset for Dutch ERC that comprises conversations scraped from Twitter.

The remainder of this paper is organized as follows. Section~\ref{sec2} first introduces the interdisciplinary field of emotion analysis, which is followed by an in-depth survey on ERC, including an overview of (i) the existing resources, (ii) the state-of-the-art machine-learning approaches, and (iii) studies at the intersection of ERC and customer service. In its turn, Section~\ref{sec3} gives a detailed description of our corpus creation by outlining the data collection process and the fine-grained annotation scheme that we designed to model emotion trajectories. The section concludes with the results of our inter-annotator agreement study and gives some suggestions on aggregating emotion labels into emotion clusters. The resulting corpus is analyzed in more detail in Section~\ref{sec4}, where we first study emotions as static attributes in isolated tweets and then (re)consider them as dynamic attributes as part of an emotion trajectory. Section~\ref{sec5} first provides a thorough discussion on the advantages and limitations of our proposed resource, which is followed by an outlook on the predictive modelling tasks and open research questions to which EmoTwiCS can be applied. Finally, Section~\ref{sec6} concludes this paper and gives some final remarks.

\section{Related research} \label{sec2}
This part is dedicated to the related research on emotion analysis and, more specifically, emotion analysis applied to conversational data (also known as ERC). While Section~\ref{EA} details the emergence of emotion analysis as an interdisciplinary research field, Section~\ref{ERC} adopts the perspective of the NLP community and zooms in on ERC. This latter part gives a detailed description of the available resources for ERC (see Section~\ref{resources}), the different machine learning approaches designed to tackle this task (see Section~\ref{ML}), and the existing research on ERC in the field of customer service (see Section~\ref{ERC-CS}).

\subsection{The advent of emotion analysis}\label{EA}

Around the shift of the century, emotions and their role in human-human or human-agent interactions started to gain a lot of interest from affective computing~\citep{Picard97}, a novel field in computer science. This emerging and interdisciplinary branch of research combines insights from disciplines such as psychology, cognitive science, social science, biomedical engineering and computer science. To recognize, process and simulate human emotions, researchers from affective computing mostly focus on physiological signals~\citep{Shu2018} and paralinguistic information (contained in, e.g., facial expressions~\citep{Li2020}, body gestures~\citep{Noroozi2021}, speech~\citep{Schuller2013,Schuller2020}). Even though the field conducts most of its research on video or audio recordings, textual transcriptions of the interactions are added in some cases~\citep{Busso2008,McKeown2012}.

In contrast, the NLP community has traditionally mostly centered its attention on information contained in textual data (e.g., reviews, social media posts, news articles). Sentiment analysis became popular in the early 2000s, largely due to the inception of social media and the corresponding rise in the volume of texts available on the Web~\citep{Liu2012}. Over the years, sentiment analysis evolved from the detection of mere polarity labels (\textit{positive}, \textit{negative}, \textit{neutral}) to (i) the identification of real-valued sentiment scores~\citep{pang-lee-2005}, (ii) the analysis of sentiment expressed towards certain aspects or features of entities (aspect-based sentiment analysis)~\citep{Pontiki2016}, (iii) the modelling of implicit sentiment~\citep{VandeKauter2015}, and (iv) the detection of more fine-grained emotions instead of polarities~\citep{Mohammad2018}. As the NLP community has only recently gained interest in emotions, many practical, theoretical and methodological issues remain to be addressed~\citep{Clavel2016,Poria2020}. This paper focuses on one of these unsolved subtasks, namely the detection of emotions and their trajectories in customer service dialogues.

\subsection{Emotion recognition in conversations (ERC)}\label{ERC}

Emotion recognition in conversations, a subfield of emotion analysis, has only recently attracted the attention of the NLP community, due to (i) the growing number of conversational data on the Web, (ii) the increased capabilities of natural language understanding (NLU) systems caused by recent developments in deep learning, and (iii) the application potential of this field to dialogue systems~\citep{Poria2019}. In this section, we first give an overview of the existing resources for the ERC task (see Section~\ref{resources}). We then continue to describe the different state-of-the-art machine learning approaches designed to tackle ERC (see Section~\ref{ML}). Finally, we conclude this overview by focusing on the intersection of ERC and customer service (see Section~\ref{ERC-CS}).

\subsubsection{Resources for ERC}\label{resources}

To our knowledge, there currently exist only a handful of publicly available corpora for the task of text-based ERC designed by the NLP community: DialyDialog~\citep{Li2017}, EmoryNLP~\citep{Zahiri2018}, EmotionLines~\citep{Chen2018}, EmoContext~\citep{Chatterjee2019}, and MELD~\citep{Poria2019b}. EmoryNLP, EmotionLines and MELD are built on the subtitles of the TV show \textit{Friends},\footnote{Even though MELD contains textual conversations, the corpus essentially extends the EmotionLines dataset to the multimodal domain, thus also including audio and video data. This extension mirrors a recent and broader trend in the field of NLP, namely the integration of multimodality.} but the EmotionLines corpus also holds a second dataset with human-to-human chat logs from Facebook Messenger. In its turn, DailyDialog accommodates dialogues from English learner websites and EmoContext comprises dialogues of three utterances between a human user and a conversational agent. The emotions in all five previously mentioned corpora are annotated along small taxonomies of three to seven emotion categories (often inspired on~\citet{Ekman1992}'s six basic emotions).

Two other frequently used resources for text-based ERC come from the field of affective computing and are called IEMOCAP~\citep{Busso2008} and SEMAINE~\citep{McKeown2012}. Both corpora are multimodal, containing not only audiovisual data, but also textual transcriptions of the dialogues. In contrast to the former resources created by the NLP community, these two datasets are more extensively annotated along both categorical and dimensional frameworks. The IEMOCAP dataset had originally been annotated along ten categories (later aggregated to six categories) and three five-point scale dimensions (\textit{valence}, \textit{activation}/\textit{arousal}, and \textit{dominance}). The SEMAINE dataset received continuous annotations (viz. $[-1, 1]$) for nine dimensions that are split in (i) five core dimensions (\textit{valence}, \textit{activation}/\textit{arousal}, \textit{power}/\textit{dominance}, \textit{anticipation}/\textit{expectation}, and \textit{emotional intensity}) and (ii) four additional dimensions selected out of a set of 27 optional rating dimensions (see annotation procedure in~\citet{McKeown2010}). \citet{McKeown2010} further provide utilities to convert dimensional annotations (including annotations of basic emotions) into categorical labels given some thresholds.

In the last couple of years, there have been a number of efforts in the NLP community on extending ERC resources with secondary information. For example, \citet{Bothe2020} introduce the EDA corpus in which they automatically label dialogue acts in the IEMOCAP and MELD datasets via a neural ensemble annotation process. Upon analyzing the resulting dataset, they find specific relations between emotions and dialogue acts (e.g., dialogue act \textit{accept}/\textit{agree} occurs frequently with \textit{joy)}. Moreover, \citet{Poria2021} extend the IEMOCAP and DailyDialog datasets with cause annotations, thus creating RECCON, the first resource for recognizing emotion causes in conversations. The authors also introduce two new challenging subtasks on their dataset, namely (i) causal span extraction and (ii) causal emotion entailment in a conversational setting.

All currently mentioned resources are in English and contain open-domain conversations. The dataset introduced in this paper focuses on conversations in Dutch, although the identified annotation tasks can be readily transferred to similar data in other languages as well. However, there already exists one very small publicly released dataset for the task of ERC in Dutch. The dataset is called deLearyous~\citep{Vaassen2012} and holds 11 textual dialogues that are grounded in the same event/scenario (namely, ``parking facilities are no longer free''). The dialogues were collected through Wizard of Oz experiments in which one participant pretended to be a manager, while the other one assumed the role of an employee. In contrast to most resources for emotion detection, the emotions in deLearyous are annotated along a dimensional framework for interpersonal communication, i.e. the Interpersonal Circumplex or Leary's Rose~\citep{Leary1957}. Emotions are rated on two orthogonal axes of which the horizontal one represents the degree of power or control, while the horizontal axis portrays the degree of agreeableness or affiliation. Given the corpus' limited size, low IAA scores and fixed grounding in the same scenario, we believe it is less suitable for more general tasks, especially beyond ERC.

\subsubsection{Machine learning approaches to ERC}\label{ML}

To give interested readers an idea of what types of machine learning methodologies can be applied to EmoTwiCS in future research, we provide an overview of existing machine learning systems designed to tackle ERC.
While earlier work on emotion detection included lexicon-based and feature-based machine learning approaches~\citep{Canales2014}, nowadays state-of-the-art results are achieved via deep learning systems. In contrast to vanilla emotion detection on isolated fragments of texts, ERC requires the additional modelling of factors such as the conversational context, the temporal order of turns, and interlocutor-specific information~\citep{Poria2019}. There are currently two competitive approaches to address ERC: either (i) the problem is framed as a sequence labelling task or (ii) the problem is defined as predicting for each timestep $t$ in the conversation the emotion $e_t$ at utterance $u_t$, given the preceding utterances ($u_{<t}$). In some variants, however, this latter task is redefined as taking both the preceding ($u_{<t}$) and future ($u_{>t}$) utterances into account (see, e.g.,~\citet{Majumder2019}). In what follows, we first describe related research on the latter task definition, thereafter returning to the former approach.

The first attempts to create context-aware representations rely on recurrent models, in which the current inputs are combined with the models' state containing information from past inputs to obtain an updated state. \citet{Poria2017} propose a long short-term memory-based network (LSTM) to extract contextual features; \citet{Hazarika2018} introduce conversational memory networks (CMNs); \citet{Majumder2019} implement other variants of the vanilla recurrent neural networks (DialogueRNNs). The latter two additionally extend their models with attention mechanisms. \citet{Li2022} propose a compact and parameter-efficient alternative by introducing BiERU, a bidirectional emotional recurrent unit that consists of a generalized neural tensor block (GNTB) to model context compositionality and a two-channel feature extractor (TFE) to extract emotional features. 
Another recurrence-based approach to ERC is called COSMIC~\citep{Ghosal2020}, a knowledge-based model that is related to DialogueRNN in its network structure and that adds information about, e.g., causal relations, mental states, and actions to improve performance. Even though these recurrence-based models can in theory handle infinitely long sequences, in practice long-term contextual information is not always propagated due to vanishing gradients and practical limitations in recurrent depth when applying backpropagation-through-time~\citep{Pascanu2013,Deboom2019}.

To mitigate these shortcomings, graph-based and transformer-based models are applied to the task of ERC. \citet{Ghosal2019} introduce a dialogue graph convolutional network (DialogueGCN) that utilizes self- and inter-party dependency information to model the conversational context. Furthermore, DAG-ERC~\citep{Shen2021} encodes utterances with a directed acyclic graph (DAG), thus combining the strengths of graph-based and recurrence-based models in terms of the information flow between long-distance and nearby context. A transformer-based approach is introduced by \citet{Lee2022} who combine in their CoMPM model the speaker's pre-trained memory with external knowledge from RoBERTa. These graph- and transformer-based models also often make use of external commonsense knowledge. As such, \citet{Zhong2019} present the knowledge-enriched transformer (KET) that interprets utterances via hierarchical self-attention and external commonsense knowledge.  
TODKAT (topic-driven knowledge-aware tranformer)~\citep{Zhu2021} is another model that fuses information from a topic-augmented language model with commonsense information extracted from external knowledge bases into a transformer-based encoder-decoder architecture. Furthermore, \citet{Li2021} consider a psychological knowledge-aware interaction graph transformer network (SKAIG) and rely on external knowledge to construct edge representations. Finally, the state-of-the-art knowledge-based ERC model is called SKIER~\citep{Li2023}, a symbolic knowledge integrated model that explicitly models the discourse relations between utterances and integrates ConceptNet~\citep{Speer2017} and SenticNet~\citep{Cambria2022} as commonsense knowledge bases.

While the previously mentioned models predict the distribution of emotions independently, the problem of ERC can also be framed as a sequence labelling task in which the globally best set of emotions is chosen for the entire conversation. By formulating ERC as a sequence labelling task, \citet{Wang2020} hope to leverage emotional consistency (which is often observed in conversations) to predict more reasonable distributions of emotion tags. They propose CESTa~\citep{Wang2020}, a contextualized emotion sequence tagging method. CESTa consists of a global context encoder (tranformer) and an individual context encoder (LSTM) which are used to learn inter-speaker and self dependencies, respectively. The output of the two encoders is concatenated in the final conditional random field (CRF) layer that makes predictions for all utterances in the conversation. In their turn, \citet{Guibon2021} are, to our knowledge, the firsts to transfer ERC from a supervised learning task to a few-shot learning sequence labelling problem. They propose ProtoSeq~\citep{Guibon2021}, a method that extends Prototypical Networks to incorporate contextual information and to consider dependencies between emotion labels. Finally, before concluding this section, we would like to emphasize that depending on the intended end application, not all the presented approaches to ERC are equally suitable. If the end application requires real-time ERC (e.g., ERC in conversational agents), then models that (i) in their architecture leverage information from future utterances or that (ii) perform sequence labelling are not appropriate. The same systems can, however, be used once a conversation has terminated to give, e.g., insights in the overall quality of customer service.

\subsubsection{The intersection of ERC and customer service}\label{ERC-CS}

As previously illustrated, most research on textual ERC focuses on clean open-domain conversations. Such resources/systems are, however, not tailored towards domain-specific real-world conversations that are not only restricted in their topics, but generally contain more noisy data and imbalanced data distributions. In this section, we present a number of studies that operate on the intersection of ERC and customer service.

One of the earliest studies in this field is introduced by \citet{Herzig2016} and applies emotion detection to customer support dialogues on social media (Twitter). The authors define two multi-label classification tasks: (i) emotion prediction on customer turns and (ii) detection of emotional techniques applied by the operating agent. To tackle both tasks, they create textual features (such as n-grams, NRC lexicon features, punctuation, emoticons) and dialogue features (such as dialogue topic, turn number, emotions/emotional techniques predicted in the previous turn, response time). \citet{Herzig2016} propose two architectures into which the features are fed: a support vector machine (SVM) and an SVM combined with a hidden Markov model (SVM-HMM). \citet{Herzig2016}'s work was pioneering at the time in the sense that they were the firsts to (i) investigate emotion detection in the context of social media conversations, to (ii) introduce dialogue features, and to (iii) predict the emotional techniques applied by operating agents. Given their well thought through innovations, our annotation framework builds upon their research (see Section~\ref{annot}).

Besides~\citet{Herzig2016}, there exists, to our knowledge, relatively little research on the application of ERC to the domain of customer service. In what follows, we give a brief overview of four other studies that focus on this specific problem. First, \citet{Mundra2017} use a CRF and neural network to predict emotions (viz.~8 distinct classes) in textual conversations from customer care contact centers. Second, \citet{Yom-Tov2018} propose a lexicon-based model to assess what they call `customer emotions' (but which are in fact mere sentiment scores, ranging from 1 to 5) in spontaneous web-based customer service interactions. Their analysis studies changes in customer sentiment during the interactions and links these changes to service quality evaluations. Third, \citet{Maslowski2017} create a hybrid machine learning system to detect interaction problems in French dialogues between a human and a virtual adviser that take place in a real-world application (namely, the chatbot of the French energy supplier EDF). In their paper, interaction problems are modelled as the expression of the user's opinions/emotions towards the interaction. Finally, \citet{Guibon2021} are the firsts to perform ERC through few-shot learning (see supra for more implementation details). Their approach is applied to the English DailyDialog dataset and a private corpus of French customer service dialogues from a live chat support.

\section{Creation of the EmoTwiCS Corpus} \label{sec3}
This section describes the construction of our Dutch Twitter corpus to detect fine-grained emotion trajectories in customer service dialogues. While Section~\ref{collect} provides more details on the Twitter conversations that we collected, Section~\ref{annot} introduces the guidelines that we designed to annotate general conversation characteristics, fine-grained customer emotions, events causing such customer emotions, and operator response strategies. Finally, Section~\ref{iaa} gives an overview of the inter-annotator agreement scores that we obtained for our four annotators across the different annotation layers on a sample of 400 conversations. As for the customer emotions, we further analyze their respective frequencies, the agreement scores for each individual emotion category, and the Jaccard similarity coefficients between different pairs of emotions. Based on these results, we propose a number of clusters to group similar emotion categories.

Before continuing our story, we briefly discuss some frequently used terminology. In this paper, we often mention terms such as \textit{tweet}, \textit{utterance} and \textit{turn}. An \textit{utterance} refers to a single message that is posted by a dialogue participant. As we are specifically working on Twitter data, we decided to consider an utterance to be a single \textit{tweet}. Both words are therefore used interchangeably throughout this paper. With the term \textit{turn}, we consider a single or any number of consecutive posts that are made by the same party (i.e., customer or operating agent) without being interrupted by the other interlocutor. This way, we adopt a similar approach as~\citet{Herzig2016} that ties in with the definitions of turn and turn-taking as they were originally proposed by~\citet{Sacks1974} and re-iterated in~\citet{Jurafsky2023-chap}.

\subsection{Data collection} \label{collect}
In prior research, we collected a multilingual conversational corpus for the domain of customer service by means of Twitter's API~\citep[see][]{Hadifar2021}. The tweets were crawled between May and October 2020 on pages of companies active in the sectors of telecommunication, public transportation or the airline industry. The resulting corpus was preprocessed by removing conversations conducted in undefined languages or conversations containing less than one interaction.\footnote{Throughout our research, we used the \textit{polyglot} natural language pipeline to automatically detect languages (see \url{https://github.com/aboSamoor/polyglot}).} The latter often occurred when tweets were removed by users or the conversation continued in a private Twitter channel.

For the intended analysis on emotion trajectories, we decided to focus on monolingual data first, and chose Dutch as the target language for the paper. To this end, we only retained conversations from firms active in Flanders, the Dutch-speaking community of Belgium. The following companies were incorporated in our selection:
\begin{itemize}
    \item \textbf{Telecommunication}: BASE, Mobile Vikings, Orange, Proximus, Scarlet, Telenet 
    \item \textbf{Public transportation}: De Lijn, NMBS
    \item \textbf{Airline industry}: Brussels Airlines, Brussels Airport, TUI fly
\end{itemize}
In a next step, we checked the number of interlocutors and we only included conversations that were held between a single customer and the company's Twitter account, which we refer to as dialogues. We also removed conversations that were not in Dutch or that code-switched between Dutch and another language. The resulting dataset contains 9,489 annotated Dutch dialogues which are, in turn, made up out of 12,715 customer utterances and 13,067 operator utterances. Table~\ref{tab:sect-conv} shows the distribution of these dialogues (in number of conversations, tweets and turns) across the different economic sectors. From the table, we can deduce that
only 3.5\% 
of all customer utterances are part of a turn with two or more tweets, while the number of operator utterances that come from a turn with two or more tweets is a lot higher (10.7\%).

\begin{table}[h!]
\caption{The number of annotated conversations, customer tweets, customer turns, operator tweets and operator turns for the three economic sectors in our corpus.}
\label{tab:sect-conv}
\centering
\begin{tabular}{lrrrrr}
\toprule 
Sector & $|\text{Convs.}|$ & \multicolumn{1}{l}{$|\text{Cust. Tweets}|$} &
\multicolumn{1}{l}{$|\text{Cust. Turns}|$} & \multicolumn{1}{r}{$|\text{Oper. Tweets}|$} & \multicolumn{1}{r}{$|\text{Oper. Turns}|$}\\
\midrule  
\midrule
Telecommunication & 5,647 & 7,820 & 7,541 & 8,858 & 7,687 \\  
Public transportation & 2,762 & 3,866 & 3,753 & 4,138 & 3,944 \\
Airline industry & 1,080 & 1,486 & 1,421 & 1,632 & 1,436 \\
\midrule 
All sectors & 9,489   &   13,172  & 12,715 & 14,628 & 13,067 \\
\bottomrule
\end{tabular}
\end{table}

\begin{figure}[t!]
    \makebox[\textwidth][c]{\includegraphics[width=0.9\textwidth]{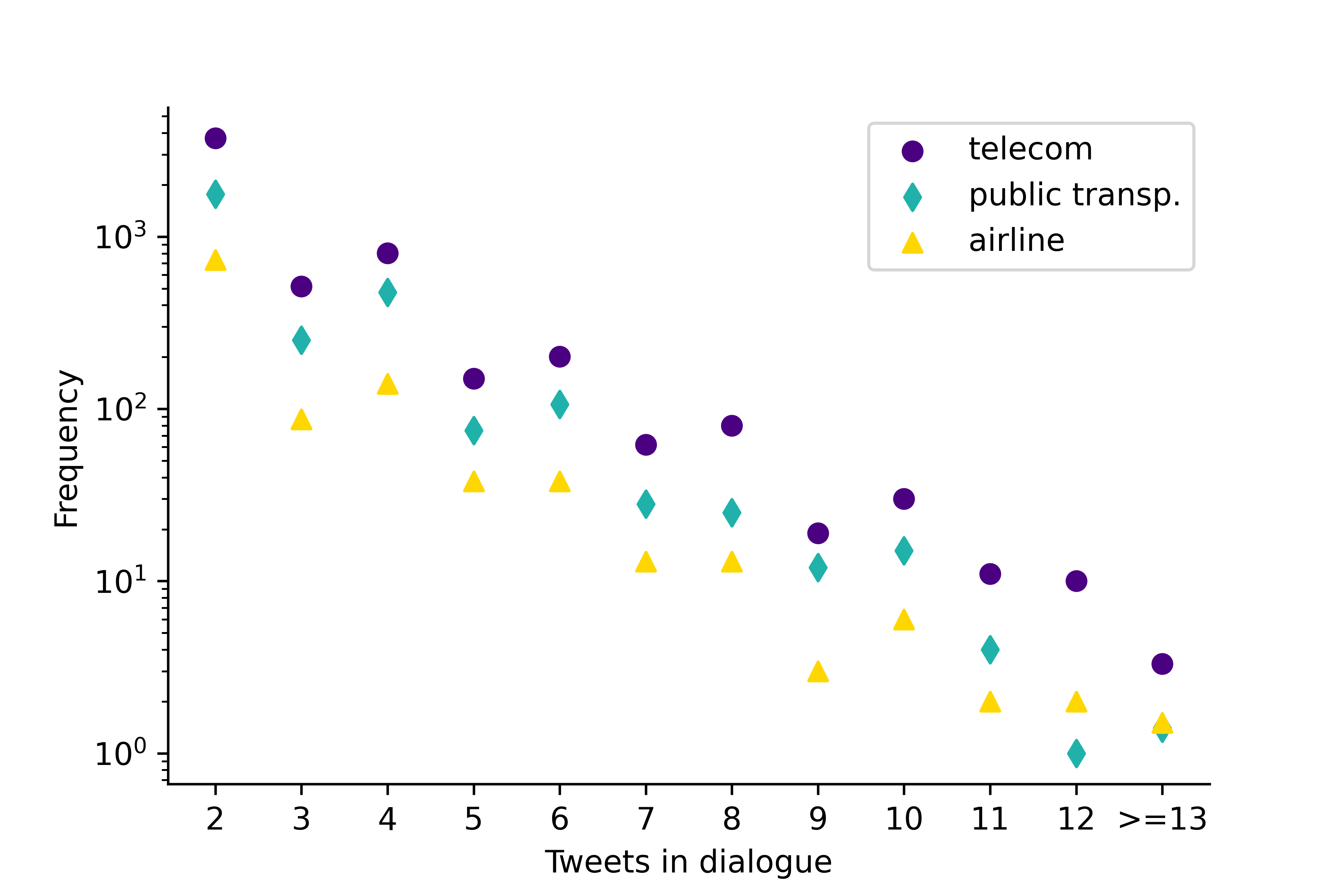}}
    \caption{Frequency of the dialogue lengths for the three sectors on a log-linear scale.}
    \label{fig:length_conv}
\end{figure}

Figure~\ref{fig:length_conv} further gives an idea of the distribution of conversation lengths (in number of tweets) for the different economic sectors. Very few conversations (7.8\%) contain six utterances or more, whereas a large fraction (65.6\% in total) only has a single customer tweet followed by a single operator tweet. The percentage of conversations with at least two customer turns with an operator turn in between is 23.0\%.  Given the research topic of customer emotion trajectories as proposed in this paper, these are of particular interest in our analysis (see Section~\ref{emo-traj}). For a possible explanation as to why most dialogues are relatively short, we refer interested readers to Section~\ref{disc1}.

\subsection{Annotation framework} \label{annot}

We designed a novel annotation scheme for the task of modelling fine-grained emotion trajectories throughout customer service dialogues. Our interdisciplinary framework is based on research from various research fields such as marketing~\citep{Simon2013}, psychology~\citep{Mehrabian1974}, and emotion analysis in NLP~\citep{Xia-Ding2019,Demszky2020}. By combining insights from these fields, we contribute to existing research in the following ways.
To our knowledge, most research dealing with ERC focuses on open-domain conversations~\citep[see, e.g.,][]{Li2017,Chen2018} in which interlocutors' emotions are often labelled along simple taxonomies that are, in many cases, variants of Ekman's model~\citep{Poria2019}. We propose, however, a slightly different approach that builds on a study from~\citet{Herzig2016} and that is more tailored towards domain-specific dialogues for which one of the interlocutors is restricted in its responses by, e.g., company policies. Moreover, while emotion cause identification has recently gained attention in the field of emotion analysis, it is still largely underexplored in conversational settings~\citep{Poria2021}. Through the proposed framework, we lay a foundation for this task in customer service dialogues and incorporate prior findings from marketing research~\citep{Simon2013} into our cause annotations.

\begin{figure}[ht!]
    \makebox[\textwidth][c]{\includegraphics[width=0.8\textwidth]{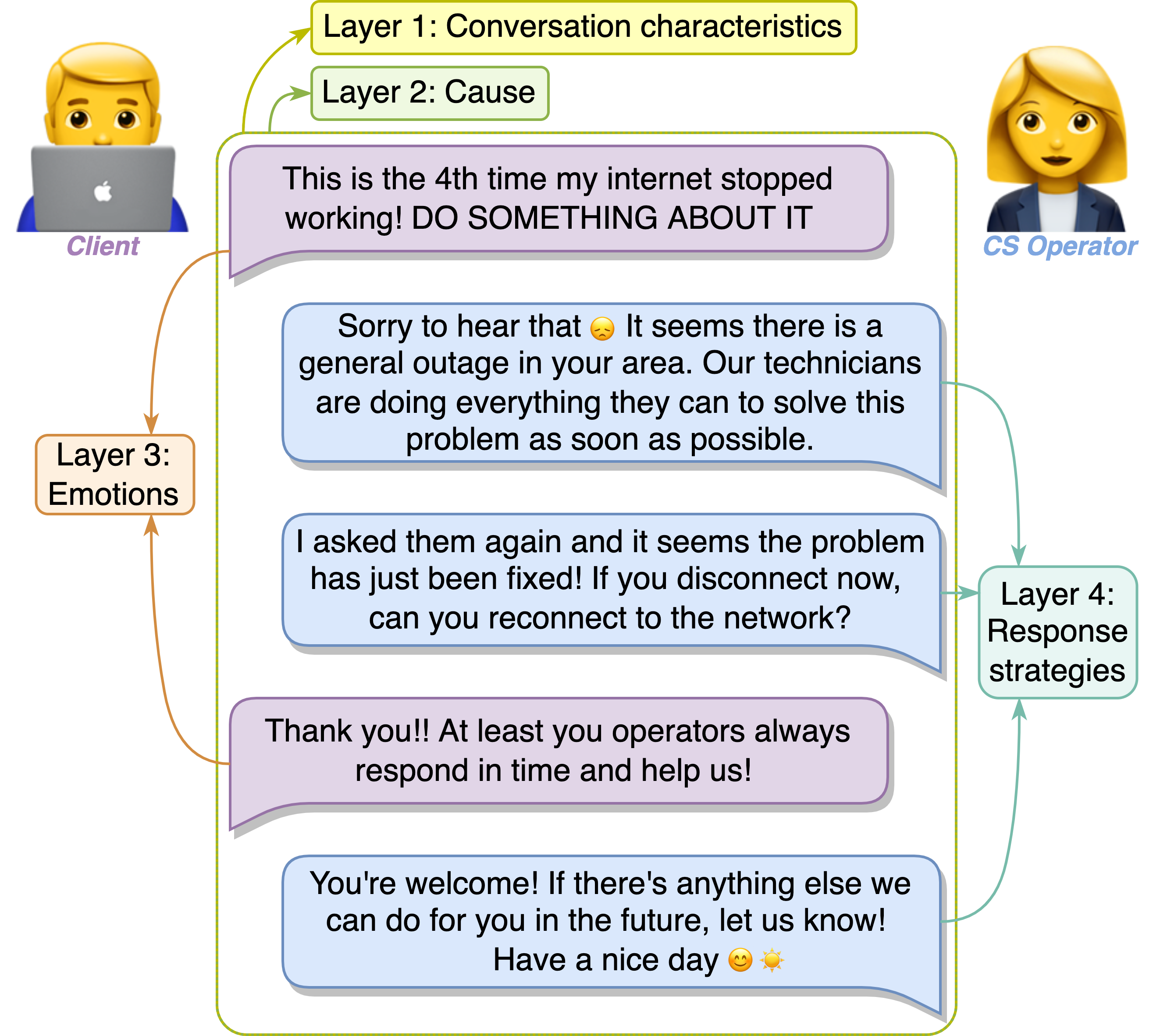}}
    \caption{Example of a customer service dialogue with corresponding illustrations of the 4 annotation layers and the different levels at which they should be annotated.}
    \label{fig:ex_conv}
\end{figure}

Our proposed annotation scheme consists of four layers, namely (i) \textbf{conversation characteristics}, (ii) \textbf{cause}, (iii) \textbf{emotions}, and (iv) \textbf{response strategies}. Figure~\ref{fig:ex_conv} gives an example of a customer service dialogue in which these annotation layers are clearly indicated at the different levels they are supposed to be annotated. Although we applied these guidelines on Dutch conversational data, the annotation framework is in principle language-independent. To further underscore this feature, the example dialogue provided in Figures~\ref{fig:ex_conv} and~\ref{fig:ex_annot} is in English, so that non-Dutch language users can understand it as well. In what follows, we provide a concise and comprehensive description of the different annotation layers. We refer interested readers to~\citet{Labat2020} for a detailed overview of the guidelines and some illustrative examples. As attentive readers might notice, the annotation framework described in this paper differs from the detailed version that was introduced in the technical report \citep{Labat2020}. The approach introduced in this paper transfers the level of our annotations to the tweet, turn or conversation level, so that our data analysis (as described in Section~\ref{sec4}) could be simplified. More details about the conversion procedure are given in the following subsections. Nevertheless, the original annotations remain at our disposal and will be investigated in future research. To annotate conversations, we recruited and trained four students from the department of Translation, Interpreting and Communication of Ghent University. The student workers were paid per hour, thus prioritizing data quality over data quantity. They also received two rounds of training and several feedback sessions. Finally, the student workers completed all annotation steps in INCEpTION~\citep{Inception_klie_2018}, an open-source annotation platform for making semantic annotations on texts.

\subsubsection{Layer 1: Conversation characteristics} \label{conv_char}
In the first and most global layer, named \textit{conversation characteristics}, annotators indicate at the conversation level whether (i) a given dialogue is subjective/objective in terms of the emotions a customer displays (viz. \textbf{subjectivity}) and whether (ii) a description of an event possibly causing these emotions is present or not at the start of the dialogue (viz. \textbf{presence cause}). In this respect, it is important to mention that a cause description can be present, while the customer does not explicitly display any emotions. Vice versa, a dialogue might start with overt customer emotions, while no actual reason (or cause description) is provided for them. The subjectivity and presence of cause inform annotators about the following annotation layers: only if a cause is present, annotators should proceed to annotate this cause in Layer 2. Similarly, only if there is at least one subjective customer tweet in the dialogue, annotators should continue to annotate the customer emotions in Layer 3 and the operator response strategies in Layer 4. We acknowledge that response strategies can, in principle, also be annotated in objective dialogues (i.e., dialogues without clear customer emotions), but we decided against this in order to reduce the workload of our student workers. Finally, annotators can flag the conversation as \textit{out of scope} in this first layer, which signals that the dialogue is in another language than Dutch (and therefore wrongfully passed through our preprocessing process, as described in Section~\ref{collect}) or that there is too little context to make sense of the dialogue. These conversations are subsequently removed from our corpus. In our case, 322 conversations of the 9,811 inspected conversations received an \textit{out of scope} annotation, leading to the 9,489 annotated conversations listed in Table~\ref{tab:sect-conv}.

\subsubsection{Layer 2: Cause}
If it is indicated in the conversation characteristics layer that the dialogue contains an explicit event description with the potential to trigger emotions, then this description is labelled at the conversation level. In the extensive annotation guidelines~\citep{Labat2020}, cause annotations were essentially made on the sentence level in one of the initial customer tweets. We can, however, easily transfer these annotations to the conversation level, since we restricted the number of event description annotations to a maximum of one per dialogue. We made this decision because we noticed that most dialogues contain at most one event description. 
For those rare cases where more than one cause description is present, annotators are instructed to only annotate the description that is the most direct cause of customer emotions. In case this rule cannot clearly be applied, then the event description that is most explicitly described in the dialogue should be annotated.

To design a proper labelling scheme for these event descriptions, we looked at related research in the field of marketing and consumer services. Our annotation scheme is based on \citet{Simon2013} who asked her students to report recent complaints they experienced as customers to investigate the role of empathy on consumer loyalty. She divided the submitted complaints in six categories of dissatisfaction: (i) employees’ difficulty in resolving problems and attending to consumers, (ii) lack of product quality, (iii) delays or service breakdowns, (iv) product information and website design inadequacies, (v) environmental or consumer health issues insufficiently addressed by company policies, and (vi) other causes. We used these categories as inspiration to design our own adapted event taxonomy, as the collected Twitter data did not only contain complaints, but also neutral remarks, questions, and compliments. Moreover, we noticed that some of the original categories (e.g., delays or service breakdowns) were so prominent in the data that we could easily split them up in two different categories. The resulting scheme consists of the following eight categories: (i) \textbf{employee service}, (ii) \textbf{product quality}, (iii) \textbf{delays and cancellations}, (iv) \textbf{breakdowns}, (v) \textbf{product information}, (vi) \textbf{digital design inadequacies}, (vii) \textbf{environmental and consumer health}, and (viii) \textbf{other}.

\subsubsection{Layer 3: Emotions} \label{annot_em}
There exist a large number of different categorical frameworks to annotate emotions. Two influential emotion taxonomies in NLP research are \citet{Ekman1992}, who proposes six universal basic emotions that are based on facial expressions: \textit{joy}, \textit{surprise}, \textit{anger}, \textit{fear}, \textit{disgust} and \textit{sadness}, and \citet{Plutchik1980}, who extends Ekman's schema to eight categories, adding \textit{anticipation} and \textit{trust}. Nevertheless, recently larger categorical sets of emotion labels are constructed~\citep[see, e.g.,][]{Cowen2017,rashkin2019}, as researchers acknowledge that our ability to recognize emotions is not limited to a relatively small set of basic emotions~\citep{Skerry2015}. We decided to adopt the taxonomy proposed by \citet{Demszky2020} which contains \textbf{27 emotion labels} and an additional \textbf{\textit{neutral} category}. Our decision for this framework is motivated by the fact that the authors applied a selection procedure in which they tried to maximize the range of emotions included in their taxonomy, while simultaneously minimizing their semantic overlap. \citet{Demszky2020} showed that their GoEmotions dataset (annotated along the proposed taxonomy) generalizes well to other domains and other taxonomies of emotions. By applying this framework to the domain of customer service, we want to not only gain a better understanding of which types of emotions frequently occur in the domain, but we will also be able to perform cross-domain and cross-lingual comparisons with our dataset.

Besides these categories, we also annotate emotions along a dimensional model. We therefore rely on a well-known model consisting of three orthogonal axes, namely \textbf{valence} (from displeasure to pleasure), \textbf{arousal} (from calm to excited), and \textbf{dominance} (from submissive to dominant) (henceforth: \textbf{VAD})~\citep{Osgood1957,Mehrabian1974}. While categorical approaches are often used to detect emotions, they suffer from a lack of consensus on the number of emotion labels in the frameworks. Moreover, emotion categories are often not equally distributed in the \textit{valence}-\textit{arousal} space~\citep{Buechel2017}. Our annotators are instructed to give an integer score ranging from 1 to 5 for each VAD dimension in which 1 represents low presence of a dimension and 5 represents high presence of that dimension. To further help annotators in correctly interpreting the different scores, we provided a visual aid called the Self-Assessment Manikin (SAM) introduced by \cite{Bradley1994} which illustrates the different emotional dimensions by means of five figures each.

If a dialogue is labelled as objective in the conversation characteristics layer, then no emotions have to be annotated. In this case, each customer tweet is automatically assigned the \textit{neutral} category and a score of 3 for each VAD dimension. If the dialogue is, however, annotated as subjective, emotions should be annotated at the level of individual customer tweets. A conversation is characterized as \textit{subjective} if a customer expresses either explicitly or implicitly his/her emotions in at least one tweet. This implies that other customer tweets in the dialogue can still be objective. In the latter case, the objective tweets are annotated with \textit{neutral} and a score of 3 for each VAD dimension. The emotions expressed in subjective tweets are, on the other hand, annotated with the 27 emotion labels and corresponding VAD scores. The number of annotations per tweet is not restricted and each annotation with an emotion category receives corresponding VAD scores. In the original guidelines, emotion annotations were often made on the word level, implying that several annotations with a same label/score for VAD dimensions could be made on a tweet. In this paper, we transfer the word-level annotations to the tweet level, thus converting identical labels or scores on one of the VAD dimensions to a single label or score, respectively.

We noticed that after converting the annotations to the tweet level, most tweets have only one score per VAD dimension. 
For those cases in which a tweet contains more than one score for an emotional dimension, the average of the set of these scores is taken and rounded to the nearest integer. In case the average lies exactly half-way between two integers, but the average is not an integer itself (i.e., it ends on .5), then the following heuristic is applied: The average is rounded off to the floor integer if the digit before the decimal separator equals 1 or 2, and rounded off to the ceil integer if this digit equals 3 or 4. This way, we opt consistently for more extreme VAD scores.

We acknowledge that depending upon the goal of the end application, better heuristics can be defined. For example, it might be interesting to prioritize negative \textit{valence} for detecting detractors and churn, while positive \textit{valence} is an important source of information to identify promoters of particular companies, services or products. Nevertheless, given the general scope of our dataset and its broad application potential, we decided to stick to the current version of the heuristic.

\subsubsection{Layer 4: Response strategies}
In contrast to emotion detection in open domain conversations, operators in customer service are restricted in their responses by company policies. \citet{Rafaeli1987} remark that the feelings employees show to customers, such as smiling and acting friendly, are part of their work role and can thus not be considered as indicators of well-being. Upon manually inspecting the raw data, we noticed that, in large part, operators try to remain objective while supporting customers. When operator emotions occur, they are part of a broader strategy that either mitigates the effect of negative customer emotions or re-enforces positive customer emotions. For those reasons, we decided to look at response strategies as emotional and informative strategies used to fulfill role expectations. Our taxonomy of response strategy labels includes the set of four emotional response techniques that \citet{Herzig2016} proposed and that are frequently applied by customer service operators, namely (i) \textbf{apology}, (ii) \textbf{cheerfulness}, (iii) \textbf{empathy}, and (iv) \textbf{gratitude}. We further added informative response techniques to the taxonomy, as we believe that customer emotions can also be influenced by adequately helping them. Our four additional informative response techniques are (v) \textbf{explanation}, (vi) \textbf{help offline}, (vii) \textbf{request information}, and (viii) \textbf{other}, thus leading to eight response strategies in total.

If the dialogue is labelled as subjective in the conversation characteristics layer, then the different response techniques have to be annotated at the level of operator tweets. As mentioned before, operator response strategies are also present in objective dialogues. Nevertheless, we decided to only annotate these strategies in subjective dialogues to reduce the workload of our working students, thus underscoring that the main focus of this paper resides on modelling customer emotions instead of operator response strategies. Therefore, all operator tweets in objective conversations automatically receive the label \textit{none}.

As for the subjective dialogues, annotators have to indicate which emotional or informative techniques the operator applied in each of his/her tweets. In the extensive guidelines described in~\citet{Labat2020}, response strategies are mostly annotated on the sentence level. In this paper, we transfer the response strategy annotations from the sentence level to the tweet level. Similarly to the transferring procedure in Section~\ref{annot_em}, we convert labels that are multiple times assigned to the same tweet to a single label (see Section~\ref{ex-annot} and Fig.~\ref{fig:ex_annot} for an illustration of this process).

\subsubsection{Illustration of our annotation framework}\label{ex-annot}

\begin{figure}[h!]
    \makebox[\textwidth][c]{\includegraphics[width=1\textwidth]{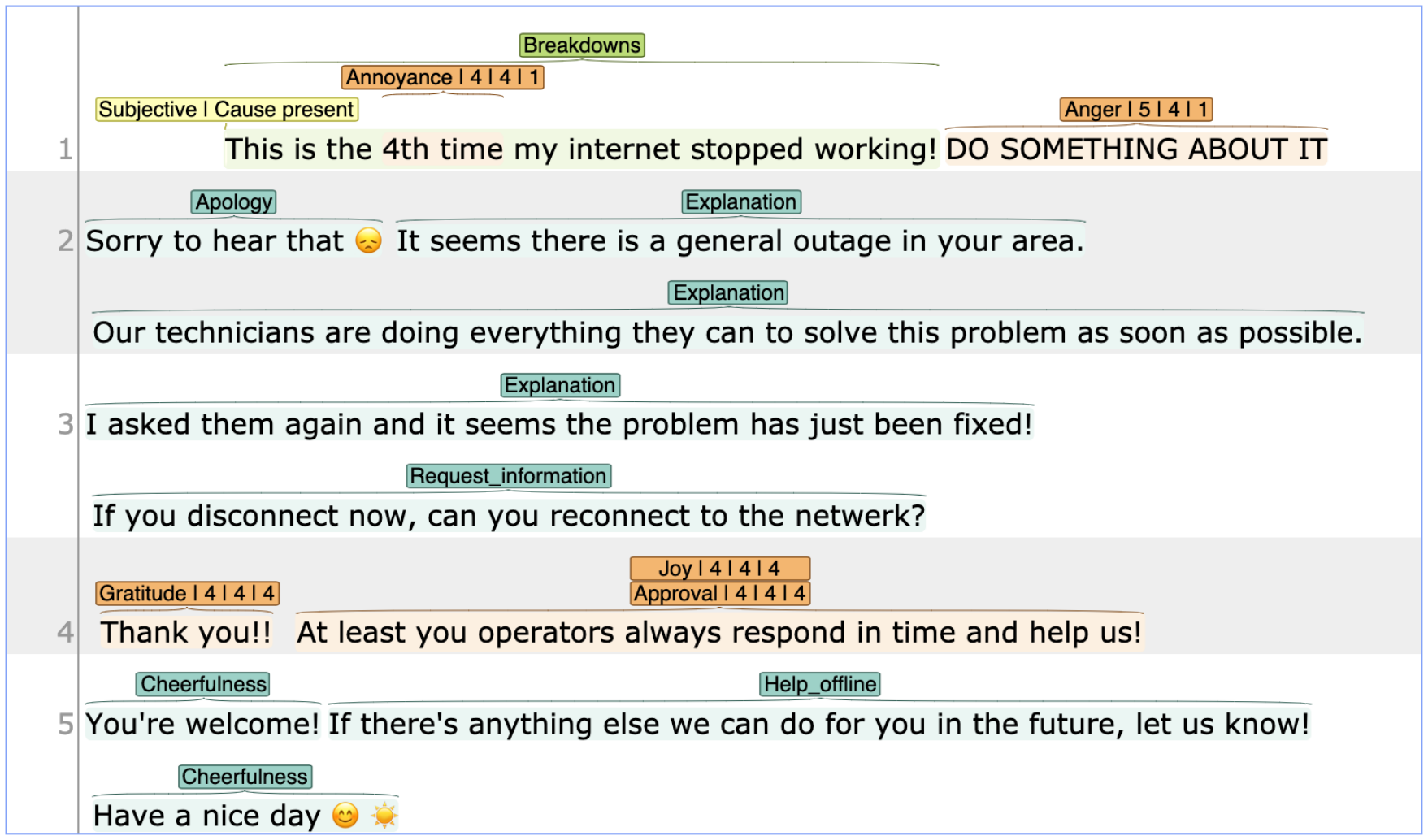}}
    \caption{Example of a dialogue that is annotated in INCEpTION for (i) \textbf{\co{conversation characteristics}}, (ii) \textbf{\ca{cause}}, (iii) \textbf{\e{emotions}}, and (iv) \textbf{\rs{response strategies}}.}
    \label{fig:ex_annot}
\end{figure}

\noindent Figure~\ref{fig:ex_annot} gives an illustration of a customer service dialogue that is fully annotated along our guidelines in INCEpTION. Each row in the figure stands for a separate tweet. The numbers in the first column represent tweet IDs, while the second column displays the text associated with each tweet. At the beginning of the dialogue, a zero-width annotation is made in yellow, indicating the conversation characteristics of the dialogue (Layer 1). As the conversation is characterized as \textit{subjective} and \textit{cause present}, all subsequent annotation layers can be applied. In the first customer tweet, a green cause annotation (Layer 2) has been made on the first sentence pointing out that this text fragment holds an event describing a \textit{breakdown}. For the purpose of this paper, we transfer the sentence level annotation to the conversation level. The same customer tweet has also received two orange emotion annotations (Layer 3). In the first orange emotion annotation, ``4th time'' hints at \textit{annoyance} and ``DO SOMETHING ABOUT IT'' conveys \textit{anger}. Note that the order of the scores in the emotion annotations of Figure~\ref{fig:ex_annot} is always alphabetical, meaning that the first number represents \textit{valence}, the second \textit{dominance}, and the last \textit{valence}. It is thus indicated that the \textit{arousal} score associated with \textit{anger} is higher than the score associated with \textit{annoyance}. When converting these \textit{arousal} scores to the tweet level with the heuristic introduced in Section~\ref{annot_em}, the resulting score becomes 5. The fourth row contains another subjective customer tweet with \textit{gratitude}, \textit{joy}, and \textit{approval}. All three emotions are associated with a VAD score of 4-4-4. As for the response strategies (Layer 4), blue annotations are made on each operator tweet. While the first and second operator tweet both carry the \textit{explanation} label, the former also holds an \textit{apology} and the latter a \textit{request information}. The final tweet in the example is characterized by \textit{cheerfulness} and \textit{help offline}.

\subsection{Inter-Annotator Agreement (IAA) Study} \label{iaa}

Annotating fine-grained emotion trajectories in conversations is not only a rather complex and subjective task, it also time-consuming. To reduce the workload of our annotators, each tweet was annotated by only one of our four annotators, implying that each annotator labelled roughly one fourth of our corpus. To gain a better understanding of the reliability of the annotations and the consistency with which our guidelines are applied, we conduct an extensive Inter-Annotator Agreement (IAA) study on 400 dialogues in our corpus (see Section~\ref{iaa-plain}). While this study gives an indication of the difficulty associated with annotating fine-grained emotion trajectories, it also forms the basis of our investigation into the ideal framework for annotating customer emotions. In Section~\ref{iaa-clust}, we look at the IAA scores obtained on the individual emotion categories (see Table~\ref{tab:iaa-emotions}) and which of these categories frequently co-occur (see Figure~\ref{fig:heatmap-jacc}). The inspections inform us for the task of clustering similar emotions. Our data analysis in Section~\ref{sec4} uses the proposed emotion clusters. 

\subsubsection{Agreement on the four layers in our annotation framework}\label{iaa-plain}

The IAA scores of the different annotation tasks and subtasks are displayed in Table~\ref{tab:iaa-layers}. All scores are calculated on a random sample of 400 dialogues. For the conversation characteristics and cause layer, this entails that we calculate IAA scores on 400 data instances each, as both layers are annotated on the conversation level. The emotion and response strategy layers contain, however, more items to label, as annotations of these layers have to be made at the level of customer and operator tweets, respectively. Depending on the annotation layer, we use a different metric to calculate agreement between annotators. Fleiss' $\kappa$ is used for those tasks where annotators have to assign mutually exclusive categories to data instances, while we applied Krippendorff's $\alpha$ in case (i) multiple, non-exclusive labels can be assigned to an instance or in case (ii) integer scores are assigned. We further specified two difference functions: Jaccard distance is used when annotators assign a set of categorical labels to each instance, while the absolute difference is used to measure annotator disagreement between exclusive ordinal annotations (namely for the \textit{valence}, \textit{arousal}, and \textit{dominance} scores). By applying Krippendorff's $\alpha$ with an absolute difference function to our integer scores, we gradually differentiate between similar or `nearby' scores (e.g., 4 vs. 5) and dissimilar or `opposite' scores (e.g., 1 vs. 5).

\begin{table}[h!]
\caption{The IAA scores (Fleiss' $\kappa$, Krippendorff's $\alpha$) on the four annotation layers.}
\label{tab:iaa-layers}
\centering
\begin{tabular}{llclr}
\toprule 
Layer & Subtask & $|\text{Items to label}|$ & Metric &
\multicolumn{1}{l}{Score}\\
\midrule 
\midrule
\multirow{2}{*}{Conversation char.}
        & Subjectivity       & 400  & $\kappa$    & 0.686 \\
        & Presence cause     & 400  & $\kappa$    & 0.680\\
\midrule 
Cause   &                    & 400  & $\kappa$    & 0.660\\
\midrule 
\multirow{4}{*}{Emotions}
        & Emotion labels & 539  & $\alpha$ (Jaccard distance) & 0.483\\
        & Valence     & 539  & $\alpha$ (Absolute difference) & 0.633\\
        & Arousal     & 539  & $\alpha$ (Absolute difference) & 0.514\\
        & Dominance   & 539  & $\alpha$ (Absolute difference) & 0.192\\
\midrule 
Response strategies &        & 596  & 
$\alpha$ (Jaccard distance) & 0.664 \\
\bottomrule
\end{tabular}
\end{table}

To properly interpret the $\kappa$ and $\alpha$ scores displayed in Table~\ref{tab:iaa-layers}, we rely on the conventions proposed by~\citet{LandisKoch1977} and~\citet{Krippendorff2004}. \citet[p.241-243]{Krippendorff2004} considers reliabilities between 0.667 $<$ $\alpha$ $<$ 0.800 for drawing tentative conclusions only, while, according to him, one can rely on values above 0.800. \citet{LandisKoch1977} regard $0.21 \leq \kappa \leq 0.40$ as fair agreement, $0.41 \leq \kappa \leq 0.60$ as moderate agreement, $0.61 \leq \kappa \leq 0.80$ as substantial agreement, and $\kappa \geq 0.81$ as almost perfect agreement. It remains, however, important to remark that most studies dealing with tasks involving the annotation of subjective and implicit data, such as emotion detection, often obtain relatively low IAA scores (see, e.g.,~\citet{Troiano2021}). Table~\ref{tab:iaa-layers} shows that we achieve substantial agreement on the conversation characteristics and cause layers. For the response strategies layer, we acquire an $\alpha$ of 0.664, close to the minimum score for drawing tentative conclusions (namely, 0.667). As for the different annotation subtasks on the emotions layer, we obtain the best result on the \textit{valence} dimension with $\alpha$ $=$ 0.633, followed by the \textit{arousal} dimension with $\alpha$ $=$ 0.514. Even though both scores are lower then 0.667, they are still relatively good considering results of related work (see, e.g.,~\citet{Antoine2014,Wood2018,DeBruyne2020}). Agreement on the \textit{dominance} dimension is, however, fairly low, but this result is in line with findings from~\citet{DeBruyne2021} who explain that for their annotation task, the difference in agreement is also largest for \textit{dominance}. Finally, our four annotators achieve a Krippendorff's  $\alpha$ of 0.483 on the emotion categories. This score is lower than the results on the \textit{valence} and \textit{arousal} dimensions, which is not that surprising, given the fact that our annotators have to choose between 28 categories when labelling customer tweets.

\subsubsection{Agreement on emotion categories and emotion clusters}\label{iaa-clust}

To gain a better understanding of how annotators agree on individual emotion categories (incl.~\textit{neutral}), we conduct another IAA study on the 400 dialogues (containing 539 tweets). The results are presented in Table~\ref{tab:iaa-emotions}. In this table, we also inspect the number of occurrences (in~\%) of (i) emotions that are assigned to tweets (Occurr. in 
\begin{table}[h!]
\caption{The IAA scores (Fleiss' $\kappa$) for individual emotion labels (including \textit{neutral}) calculated on a set of 539 customer tweets. The table further shows the number of occurrences (in \%) of these labels in (i) being used in an annotation (Occurr. in annots.) and (ii) being assigned to a tweet by at least one annotator (Occurr. in tweets).}
\label{tab:iaa-emotions}
\centering
\begin{tabular}{lrrr}
\toprule 
Emotion label & Occurr. in annots. (\%) & Occurr. in tweets (\%) &\multicolumn{1}{l}{Fleiss' $\kappa$}\\
\midrule 
\midrule
Gratitude   & 8.2 & 5.4 & 0.881\\ 
Neutral     & 40.0 & 29.0 & 0.675\\ 
Love        & 0.5 & 0.5 & 0.482\\ 
Sadness     & 1.3 & 1.4 & 0.466\\ 
Admiration  & 1.2 & 1.4 & 0.459\\ 
Anger       & 5.1 & 5.7 & 0.457\\ 
Annoyance   & 17.0 & 17.8 & 0.417\\ 
Joy         & 1.6 & 2.0 & 0.389\\ 
Amusement   & 1.4 & 1.8 & 0.334\\ 
Caring      & 0.3 & 0.5 & 0.331\\ 
Desire      & 2.6 & 3.4 & 0.331\\ 
Fear        & 0.3 & 0.5 & 0.331\\ 
Disapproval & 7.7 & 9.9 & 0.323\\ 
Disgust     & 1.2 & 1.6 & 0.278\\ 
Disappointment & 2.1 & 3.2 & 0.266\\ 
Excitement  & 0.4 & 0.6 & 0.263\\ 
Optimism    & 0.9 & 1.3 & 0.248\\ 
Confusion   & 3.6 & 5.8 & 0.237\\ 
Approval    & 1.6 & 2.7 & 0.173\\ 
Nervousness & 1.1 & 2.0 & 0.115\\ 
Relief      & 0.3 & 0.5 & 0.109\\ 
Realization & 0.5 & 1.0 & 0.099\\ 
Embarrassment & 0.0 & 0.1 & 0.0\\ 
Pride       & 0.1 & 0.3 & 0.0\\ 
Remorse     & 0.0 & 0.1 & 0.0\\ 
Surprise    & 0.2 & 0.5 & 0.0\\ 
Curiosity   & 0.7 & 1.4 & -0.004\\ 
Grief       & 0.0 & 0.0 & NA\\ 
\bottomrule
\end{tabular}
\end{table}
annots.) and of (ii) tweets that receive at least one annotation with that emotion label (Occurr. in tweets). For example, if a tweet receives three times the label \textit{annoyance} by three different annotators, then $|\text{Occurr. in annots.}| = 3$ and $|\text{Occurr. in tweets}| = 1$ for that given label. The occurrences in annotations are normalized over the total number of emotion labels used by our annotators to annotate the conversations in our IAA study (2,443 labels). The occurrences in tweets are normalized over the total number of unique emotion labels assigned to each of the 539 tweets (1,111 labels). We also calculated an average weighted Fleiss' $\kappa$ over the different emotion labels. The results of this are reported in Table~\ref{tab:iaa-emotions-avg} in the Appendix.

Frequently occurring categories seem to be linked to higher Fleiss' $\kappa$ scores and vice versa. However, this correlation is not applicable to all labels, as we achieve, for example, high $\kappa$ scores on less frequent categories such as \textit{love} and, inversely, low $\kappa$ scores on more frequent categories such as \textit{confusion}. Table~\ref{tab:iaa-emotions} further indicates that \textit{gratitude}, \textit{neutral}, \textit{love}, \textit{sadness}, \textit{admiration}, \textit{anger}, and \textit{annoyance} have the highest IAA scores on which all annotators have moderate agreement (Fleiss' $\kappa > 0.41$), while \textit{realization}, \textit{relief}, \textit{nervousness}, and \textit{approval} are categories with poor agreement (Fleiss' $\kappa < 0.21$). Some categories (viz.~\textit{embarrassment}, \textit{pride}, \textit{remorse}, \textit{surprise}, \textit{grief}) were so infrequent that we could not properly calculate agreement between annotators and the emotion \textit{grief} did not occur in the sample at all. Moreover, even though \textit{curiosity} occurred 16 times, which is more frequent than some other emotion categories, this emotion seems particularly daunting to label, as annotators did not achieve any type of agreement on it ($\kappa = -0.004$).

\begin{figure}[h!]
    \makebox[\textwidth][c]{\includegraphics[width=0.9\linewidth]{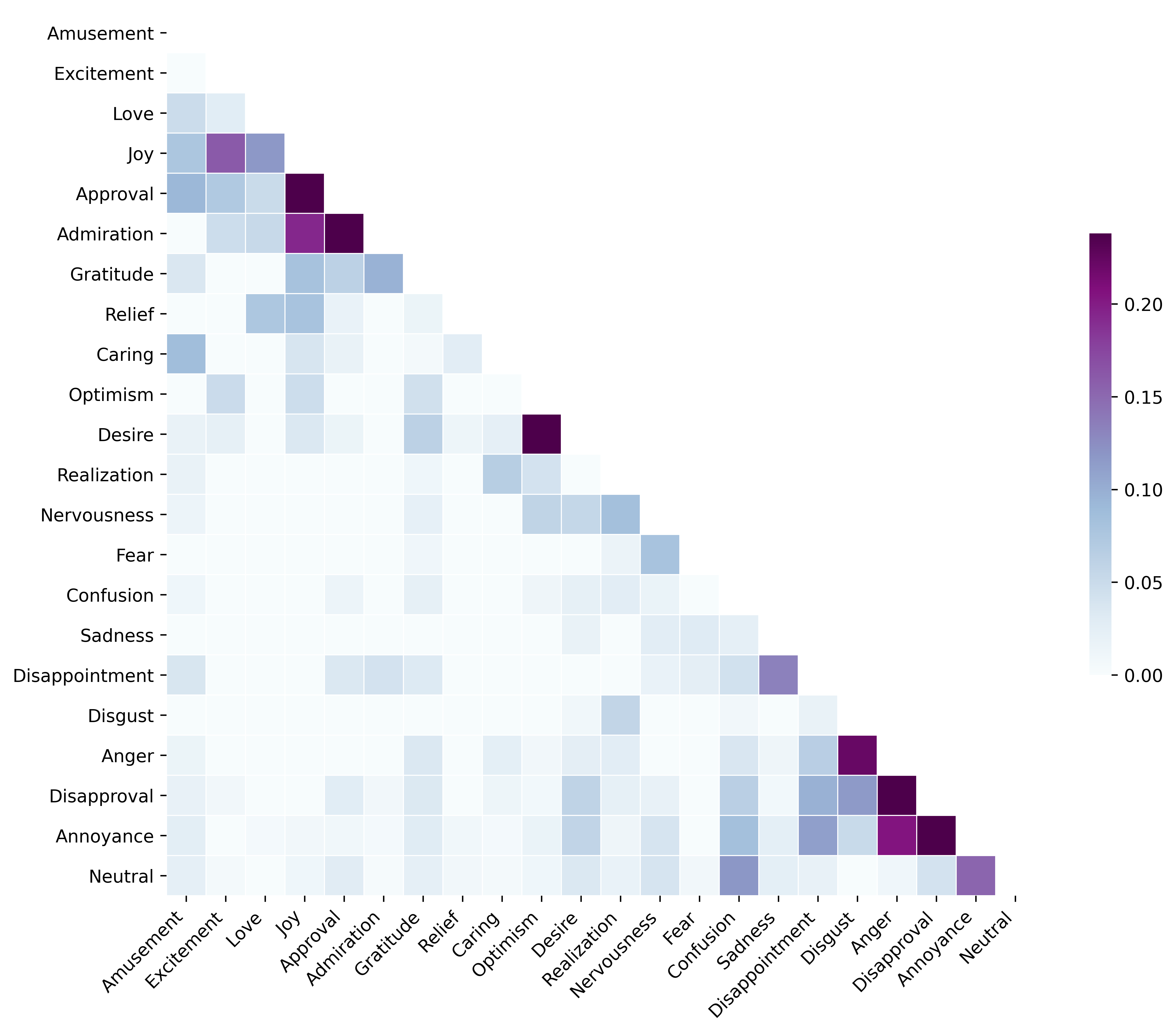}}
    \caption{The heatmap shows the Jaccard similarity coefficient between the set of tweets containing emotion $e_i$ and the set of tweets containing emotion $e_j$ for which $i$, $j$ $\in$ \textit{E}, \textit{E} is the set of emotions (incl. \textit{neutral} category), and $i$ $\neq$ $j$.}
    \label{fig:heatmap-jacc}
\end{figure}

Even though \citet{Demszky2020}'s emotion taxonomy provides interesting opportunities for cross-domain comparisons, we suspect that there is some semantic overlap between certain emotions in the domain of customer service. To gain a better idea of which types of emotions frequently co-occur in customer tweets, we plot a heatmap in Figure~\ref{fig:heatmap-jacc}. This heatmap displays the Jaccard similarity between the joined sets of emotion annotations each annotator made per customer tweet. The Jaccard similarity coefficient is calculated as the fraction of the size of the intersection over the size of the union of two sets:
\begin{gather*}
J(T_{e_i}, T_{e_j})=\frac{|T_{e_i} \cap T_{e_j}|}{|T_{e_i} \cup T_{e_j}|}.
\end{gather*}
In this formula, $T_{e_i}$ stands for the set of tweets annotated with emotion $e_i$, while $T_{e_j}$ stands for the set of tweets containing emotion $e_j$. Both $i$ and $j$ belong to $E$, the set of emotion labels (incl. \textit{neutral} category), and $i \neq j$. It is important to remark that with emotion labels, we refer to those categories that received at least 0.099 as $\kappa$ score. The higher the Jaccard similarity, the more often two emotions co-occur in customer tweets, which probably points to the fact that they have similar meanings.

On the basis of Figure~\ref{fig:heatmap-jacc} and the frequencies listed in Table~\ref{tab:iaa-emotions}, we propose a number of emotion clusters in Table~\ref{tab:emo-clusters}.\footnote{We acknowledge that multiple clusters are equally probable, but based on our informed decision, we decided to work with these clusters in the data analysis of Section~\ref{sec4}.} To not exclude the infrequent labels with poor IAA scores from Table~\ref{tab:iaa-emotions}, we add these emotion categories to the \textit{neutral} cluster. Due to this decision, the \textit{neutral} cluster can in principle co-occur with other emotion clusters. For example, if a tweet is labelled with both \textit{joy} and \textit{pride}, then these annotations are aggregated to \textit{joy} and \textit{neutral} on the cluster level.  This is not exactly what we want, as \textit{neutral} signals that emotion is absent. We therefore decided to remove the \textit{neutral} cluster if it co-occurs with other emotion clusters.
In the remainder of this paper, we continue to work with the emotion clusters as presented in Table~\ref{tab:emo-clusters}. 

By clustering emotion labels, we are able to improve the initial Krippendorff's $\alpha$ of 0.483 (on individual emotion labels) to 0.622. Note that similar improvements can be found when considering the average weighted Fleiss' $\kappa$ scores as reported in Table~\ref{tab:iaa-clusters-avg} in the Appendix. Table~\ref{tab:emo-clusters} lists the emotion clusters and their IAA score. Substantial agreement is achieved on the clusters \textit{gratitude}, \textit{neutral} and \textit{joy}, while we obtain moderate agreement on \textit{anger}, \textit{annoyance}, \textit{desire} and \textit{disappointment}. As for the residual clusters \textit{relief} and \textit{nervousness}, these are characterized by fair and low IAA, respectively. The table further shows the number of unique occurrences (in \%) of these clusters in (i) an annotation (Occurr. in annots.) and in (ii) a tweet (Occurr. in tweets). The occurrences in annotations are normalized over the total number of clusters used by our annotators to annotate the conversations in our IAA study (2,330 clusters). The occurrences in tweets are normalized over the total number of unique clusters assigned to each of the 539 tweets (869 clusters).

\begin{table}[h!]
\caption{Emotion clusters, the emotion labels that compose them, the number of occurrences (in \%) of a cluster appearing in an annotation of the annotators (Occurr. in annots.), the number of occurrences (in \%) of a cluster being at least once assigned to a tweet (Occurr. in tweets), and their IAA scores (Fleiss' $\kappa$).}
\label{tab:emo-clusters}
\centering
\begin{tabular}{llrrr}
\toprule
Emotion cluster & Emotion label(s) & Occurr. annots. (\%) & Occurr. tweets (\%) &\multicolumn{1}{l}{Fleiss' $\kappa$}\\
\midrule
\midrule
Gratitude   & Gratitude & 8.6 & 6.9 & 0.881\\ 
Neutral & Confusion, Curiosity, & 45.4 & 38.4 & 0.725 \\ 
    & Embarrassment, Neutral, \\
    & Pride, Realization, \\
    & Remorse, Surprise \\
Joy & Admiration, Amusement, & 6.2 & 6.8 & 0.693\\ 
    & Approval, Excitement,\\
    & Joy, Love \\
Anger & Anger, Disgust  & 6.5 & 7.6 & 0.535\\ 
Annoyance & Annoyance, Disapproval & 24.2 & 26.5 & 0.527\\ 
Desire & Desire, Optimism & 3.6 & 4.7 & 0.521\\ 
Disappointment & Disappointment, Sadness & 3.5 & 5.2 & 0.443\\ 
Relief & 
Caring, Relief & 0.6 & 1.0 & 0.252\\ 
Nervousness & 
Fear, Nervousness & 1.5 & 2.9 & 0.199\\ 
\bottomrule
\end{tabular}
\end{table}

\section{Data analysis} \label{sec4}
This section is dedicated to a detailed analysis on our final corpus. First, Section~\ref{emo-isol} investigates the distribution of isolated customer emotions which are expressed in emotion labels, emotion clusters, and dimensional \textit{valence}, \textit{arousal} and \textit{dominance} scores. Section~\ref{emo-traj} further zooms in on emotion trajectories by considering how emotions evolve during a conversation (see Section~\ref{emo-traj-evo}). In a next step, we investigate the link between emotions and their causes and response strategies in the context of emotion trajectories (see Section~\ref{cause-emo} and~\ref{resp-emo}, respectively).

The data analysis presented in this part is solely conducted on subjective dialogues (5,299 conversations in total), as these are the conversations that received annotations for emotions and response strategies.\footnote{One exception to this is Table~\ref{tab:freq-cause}, as it displays frequencies of cause categories in both subjective and objective conversations.} Furthermore, we include the IAA conversations in our analysis. To this end, we only selected the annotations that were made by the main author, as they overall seemed to achieve the highest IAA with the annotations made by other annotators. While Section~\ref{emo-isol} focuses on emotions in isolated tweets, Section~\ref{emo-traj} analyses emotion trajectories and considers our conversational data at the level of customer/operator turns. To move to turn level, we aggregated the annotations that were made on consecutive tweets of the same dialogue participant (i.e., belonging to the same turn), and removed classes/scores that occurred multiple times. From the resulting sets of emotion labels and clusters, we removed the category \textit{neutral} if it occurred in combination with other emotion labels/clusters. This way, we ensured to only use the \textit{neutral} category on turns void of any emotions. As for the VAD annotations, we applied the same heuristic as introduced in Section~\ref{annot_em} to ensure that all turns have only one score for each dimension.

\subsection{Emotions in isolated customer tweets} 
\label{emo-isol}

Table~\ref{tab:freq-vad} displays the distribution of VAD scores in our dataset, while Table~\ref{tab:freq-emo} does the same for emotion clusters. The frequencies in both tables are calculated on the level of customer tweets for all conversations that are labelled as subjective. Additional frequencies on turn level are available in the Appendix (see Table~\ref{tab:ext-freq-vad} and~\ref{tab:ext-freq-emo}). Frequencies of emotion labels (instead of clusters) are also shown in the latter. When considering \textit{valence} in Table~\ref{tab:freq-vad}, we notice that more than half of the customer tweets (57.4\%) are marked with a negative sentiment (V $\in \{1, 2\}$) and only 15.8\% contains a positive sentiment (V $\in \{4,5\}$). The remaining 26.8\% of these utterances are neutral (V $= 3$). These findings seem to suggest that in the context of business-to-consumer (B2C) interactions, Twitter is mostly used as a platform for handling complaints and providing customer support. For \textit{arousal}, almost all customer tweets have an average (A~$=~3$) to high (A $\in \{4, 5\}$) score on this dimension. Several explanations for this trend are possible: (i) tweets generally contain more explicitly `aroused' language such as emojis, capitalization, word lengthening, etc.; (ii) customers who actively contact firms are more aroused than those who decide to not undertake any action; and (iii) most emotions in our taxonomy are linked to medium-to-high \textit{arousal} scores (see, e.g.,~\citet{Toisoul21}).
Finally, the scores assigned to the \textit{dominance} dimension are mostly centered around the middle score of 3.

\begin{table}[h!]
\caption{Frequency (in \%) of VAD scores in tweets of subjective conversations.}
\label{tab:freq-vad}
\centering
\begin{tabular}{lrrrrr}
\toprule
 & 1 & 2 & 3 & 4 & 5 \\
\midrule
\midrule
Valence & 29.1 & 28.3 & 26.8 & 11.0 & 4.8 \\
Arousal & 0.1 & 3.2 & 39.9 & 50.4 & 6.5 \\
Dominance & 6.2 & 19.8 & 43.7 & 28.4 & 1.9\\
\bottomrule
\end{tabular}
\end{table}

\begin{table}[h!]
\caption{Frequency (in \%) of emotion clusters in tweets of subjective conversations.}
\label{tab:freq-emo}
\centering
\begin{tabular}{lr}
\toprule
Emotion cluster & \multicolumn{1}{l}{Freq.}\\
\midrule
\midrule
Annoyance & 37.9\\
Neutral & 20.3\\
Gratitude & 13.5\\
Joy & 7.8\\
Anger & 7.8\\
Disappointment & 5.3\\
Desire & 4.1\\
Nervousness & 2.0\\
Relief & 1.3\\
\bottomrule
\end{tabular}
\end{table}

For emotion clusters, we see that the most frequent emotion is \textit{annoyance}, as it occurs in 37.9\% of the customer utterances present in subjective conversations. \textit{Neutral} still occurs in one fifth (20.3\%) of the customer tweets, indicating that subjective conversations do not necessarily contain emotions in each utterance. The two most frequent positively associated emotions, \textit{gratitude} and \textit{joy}, are present in 13.3\% and 7.7\% of all the cases, respectively. As for negatively associated emotions, \textit{anger} and \textit{disappointment} are linked to 7.7\% and 5.2\% of the customer utterances, respectively. Furthermore, \textit{desire} occurs in 4.0\% of all instances, while the residual categories \textit{nervousness} and \textit{relief} only make up 1.9\% and 1.2\% of the customer tweets.

\begin{figure}[ht!]
    \makebox[\textwidth][c]{\includegraphics[width=1\linewidth]{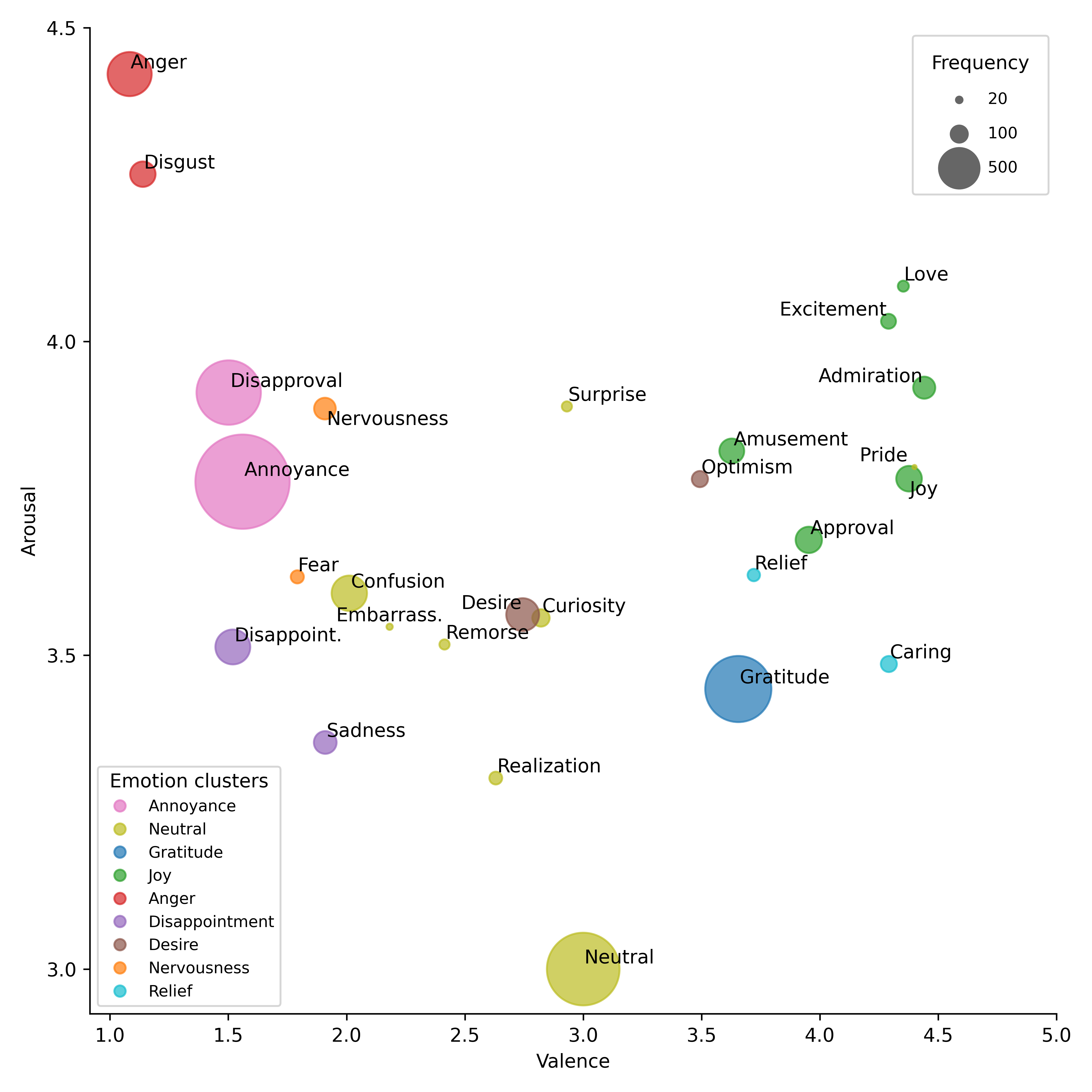}}
    \caption{Distribution of emotion labels and clusters of subjective tweets plotted in the valence-arousal space.}
    \label{fig:scat-emo}
\end{figure}

To better understand how VA scores and emotions (expressed in clusters and, by extension, emotion labels as introduced in Section~\ref{sec3}) relate to each other, we added Figure~\ref{fig:scat-emo}. We decided to only work with \textit{valence} and \textit{arousal} (thus leaving \textit{dominance} out) for two reasons. First, plotting dots in a 2D-space allows for cleaner visualizations than in a 3D-space. Second, we obtained low IAA scores on the \textit{dominance} dimension. In this figure, each emotion label is plotted in a 2D-space by averaging the \textit{valence} and \textit{arousal} scores associated with all its occurrences in tweets. The emotion labels are grouped into emotion clusters by their hue, while the size of each dot represents the frequency (on tweet level) of the emotions in our dataset. We would like to remark that in contrast to the suggestion of~\citet{Demszky2020}, desire is not necessarily a positive emotion in the context of customer service, as it occurs in both positive and negative turns. Intuitively, the positions of most emotion labels in the 2D-space make sense: more extreme \textit{valence} values (such as 1 or 5) receive higher \textit{arousal} scores. The figure clearly shows that higher \textit{valence} scores correlate positively with \textit{arousal}, while lower \textit{valence} scores correlate negatively with \textit{arousal}. This U-shaped relationship between \textit{valence} and \textit{arousal} has also been described in previous research~\citep{Warriner2013,Mattek2017}.

\subsection{Emotions trajectories in conversations}\label{emo-traj}

In the previous subsection, we analyzed customer emotions as static states in isolated tweets. We now revisit this assumption and consider them as dynamic entities at the level of customer turns in a conversational setting. In Section~\ref{emo-traj-evo}, we study (i) how \textit{valence} generally evolves from the start to the end of conversations and (ii) how emotion clusters change from one customer turn to the next. Thereafter, we explore how other factors such as causes and response strategies are linked to emotions (see Section~\ref{cause-emo} and~\ref{resp-emo}, respectively).

\subsubsection{Evolution of emotions throughout conversations}\label{emo-traj-evo}

\begin{figure}[ht!]
    \makebox[\textwidth][c]{\includegraphics[width=0.8\linewidth]{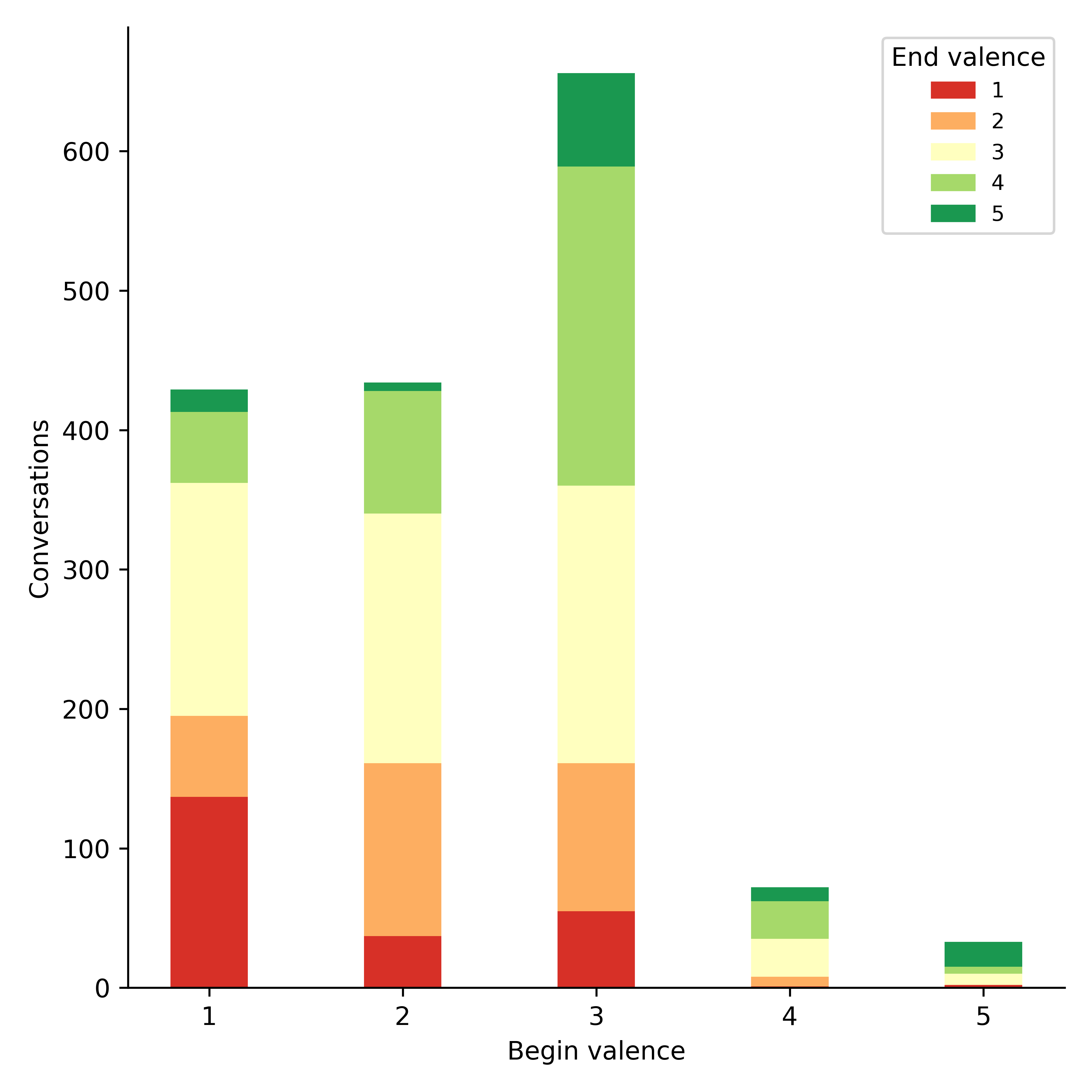}}
    \caption{Stacked bar plot showing the evolution of valence in conversations.}
    \label{fig:val-begin-end}
\end{figure}

We assume that operators at all times try to help their customers, thus transferring the emotional state of the latter towards a more positive one or attempting to retain the emotion state in a positive one. To investigate whether this is actually the case, we looked at all subjective conversations with at least two customer turns (in total 1,624 conversations) and then compared the \textit{valence} of the first customer turn to the \textit{valence} of the last one. The results of this comparison are shown in Figure~\ref{fig:val-begin-end}. The position of the stacked bars on the x-axis represents the begin \textit{valence}, while the hues in each of these stacked bars portray the end \textit{valences} for that particular begin \textit{valence}.

From Figure~\ref{fig:val-begin-end}, we see that more than half of the conversations begin with a negative \textit{valence} of 1 or 2 (53.1\%). When comparing the bars with begin \textit{valence} 1 and 2 to each other, we notice that a considerate number of conversations have not changed their \textit{valence} scores at the end of the emotion trajectory (31.9\% of convs. starting with V $= 1$; 28.6\% of convs. starting with V $= 2$). In 8.5\% of the conversations with start \textit{valence} 2, the \textit{valence} has even become more negative at the end of the emotion trajectory. However, the most frequently occurring end \textit{valence} is 3 in both cases, while positive end \textit{valences} are only obtained in a small percentage of conversations (15.6\% of convs. starting with V $= 1$; 21.7\% of convs. starting with V $= 2$).

As for the conversations that start with a neutral \textit{valence} (i.e., V $= 3$), almost half of them (45.1\%) end their emotion trajectory in a more positive \textit{valence} and almost a third of them (30.3\%) remains in the same \textit{valence} at the end of the trajectory. Nevertheless, in 24.5\% of these conversations, the end \textit{valence} has transferred to a negative one at the end of the emotion trajectory. We propose several hypotheses to explain trend: (i) the customer might not have been properly helped; (ii) the operator might have tried to assist the customer to the best of his/her capabilities, but this assistance was not enough to solve the problem or fulfill the intent of the customer; (iii) the customer initially contacted the company in a neutral tone, already feeling slightly negative (e.g., frustrated, disappointed), but towards the end of the conversation these negative emotions became more explicit. 

Finally, only a very small number of emotion trajectories begin with a positive sentiment (6.5\%), which indicates that Twitter is mainly used as a channel to handle complaints or provide customer support. Most of these conversations starting in a positive \textit{valence} have retained that positive \textit{valence} at the end of the emotion trajectories (51.4\% of convs. starting with V $= 4$; 69.7\% of convs. starting with V $= 5$), while another substantial subset ended the emotion trajectory in a neutral \textit{valence} (37.5\% of convs. starting with V $= 4$; 24.4\% of convs. starting with V $= 5$). Finally, 11.1\% of the conversations with a begin \textit{valence} of 4 ended in a negative \textit{valence}, while this percentage is 6.1\% for the conversations starting with a \textit{valence} of 5.

\begin{figure}[ht!]
    \makebox[\textwidth][c]{\includegraphics[width=1\linewidth]{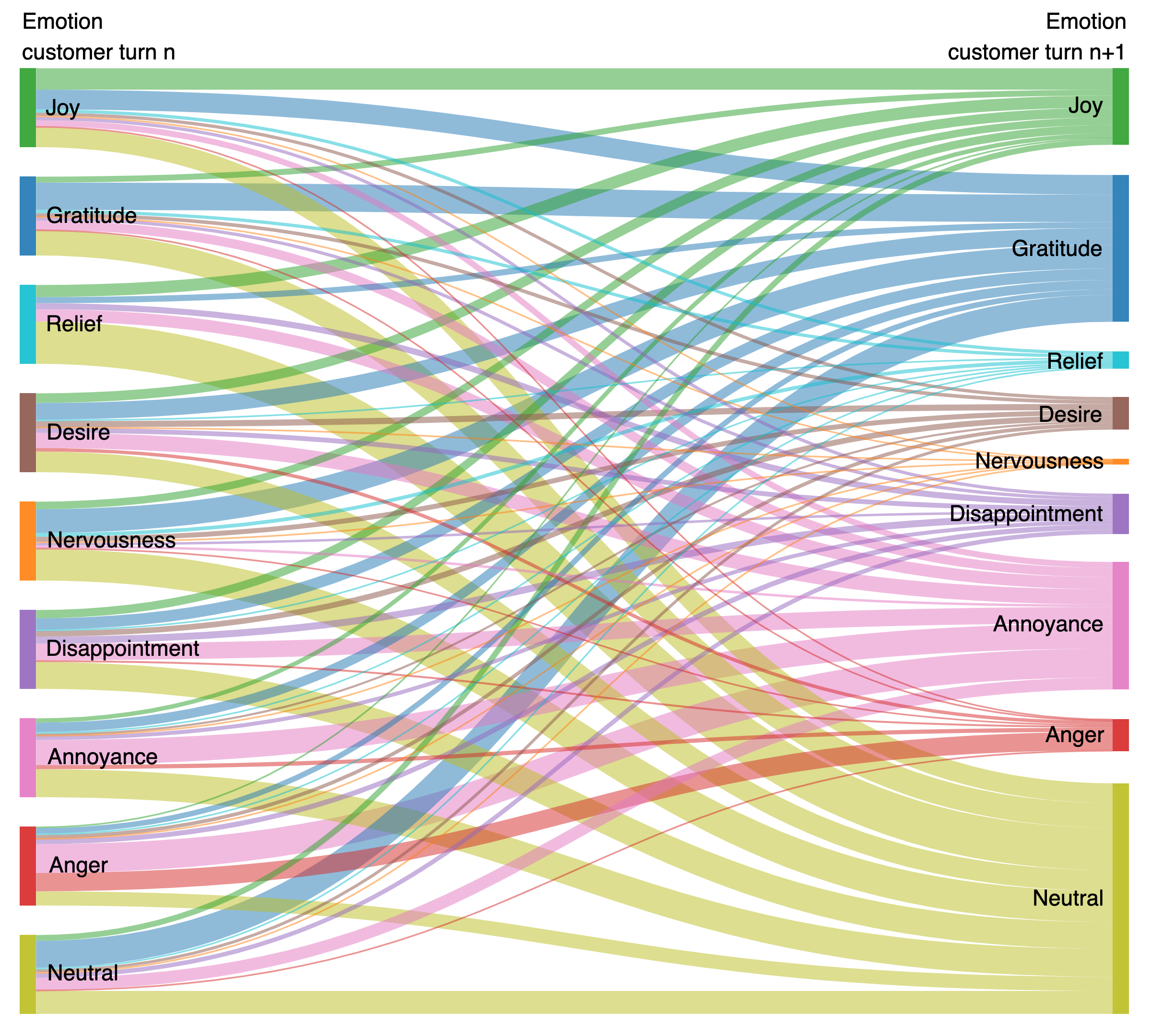}}
    \caption{Sankey diagram plotting the transitions in emotion clusters between consecutive customer turns.}
    \label{fig:emo-turn-to-turn}
\end{figure}

Although Figure~\ref{fig:val-begin-end} gives an interesting overview of how sentiment (or \textit{valence}) progresses from the beginning to the end of customer service conversations, we also want to understand how actual emotions (instead of mere sentiment) evolve from one customer turn to another. The Sankey diagram in Figure~\ref{fig:emo-turn-to-turn} plots the shifts in emotion clusters between two adjacent customer turns (separated only by a single operator turn) for all 2,500 instances in our corpus. In the diagram, we normalized the emotions at turn $n$, so that we can properly compare connections from different emotions at that position. For a given emotion cluster at position $n$, the width of the connections represents the proportion of turns that transition, while their hues corresponds to the types of emotions in the next customer turn.

All emotions seem to frequently move towards the \textit{neutral} cluster in the next customer turn. This tendency is, however, less pronounced for the two most `extreme' emotion clusters (in terms of average \textit{valence} and \textit{arousal} scores, see Fig.~\ref{fig:scat-emo}) in our corpus. The positive emotion clusters \textit{joy} and \textit{gratitude} do not change from one turn to another in 27.1\% and 34.7\% of the cases, respectively. Even though a considerate proportion of turns labelled with \textit{joy} shift to \textit{gratitude} in the next customer turn, the same does not hold up for the trajectory from \textit{gratitude} to \textit{joy}. Other clusters such as \textit{nervousness},  \textit{neutral} and \textit{desire} also regularly transition to \textit{gratitude} in the next customer turn. Figure~\ref{fig:emo-turn-to-turn} further indicates that \textit{desire} is not clearly linked to a specific sentiment. Depending on whether a customer intent is fulfilled or not, emotion trajectories with \textit{desire} can either progress to positive emotions such as \textit{gratitude} and \textit{joy} or negative emotions such as \textit{disappointment} and \textit{annoyance}. As for the negative emotion clusters, we see that trajectories that carry \textit{anger} in a specific turn often continue to carry that \textit{anger} in the next customer turn. Nevertheless, \textit{annoyance} is the negative emotion to which the negative emotions \textit{anger}, \textit{annoyance} and \textit{disappointment} most frequently flow. Finally, of these three emotions, \textit{disappointment} evolves the most often to positive emotions such as \textit{joy} and \textit{gratitude} in the next turn, followed by \textit{annoyance}. In its turn, \textit{anger} almost never shifts to positive emotions in the next customer turn.

\subsubsection{The role of causes in emotion trajectories}\label{cause-emo}

We now turn our attention to the causes in our corpus.
Remember that causes are explicit event descriptions with the potential to trigger emotions that usually occur at the beginning of the conversation. Besides providing an overview of their frequencies, we also investigate the relationship between them and the emotions in the context of emotion trajectories. Table~\ref{tab:freq-cause} describes the frequencies of the different cause labels in all 5,722 dialogues that are indicated to contain a cause description. For both the subjective and objective dialogues in this subset, \textit{breakdowns} are by far the most frequently occurring events that cause customers to contact their company on Twitter. With a frequency of 18.0\%, \textit{employee service} is the second most frequent event in the subset of subjective dialogues with a cause, while \textit{digital design inadequacies} is the second most frequent event in the objective subset with a frequency of 19.7\%. 
By comparing the frequencies of events in subjective dialogues to their frequencies in objective dialogues, we find that some of these events are clearly more present in conversations that have emotions and others are more associated with objective conversations. Causes falling under the former category are \textit{employee service}, \textit{delays and cancellations} and \textit{environmental and consumer health}. In contrast, causes that are more often linked to objective conversations are \textit{breakdowns} and \textit{design inadequacies}.

\begin{table}[h!]
\caption{Frequencies (in \%) of cause types across all conversations containing a cause.}
\label{tab:freq-cause}
\centering
\begin{tabular}{lrrr}
\toprule
Cause & \multicolumn{1}{l}{Subj. convs.} & \multicolumn{1}{l}{Obj. convs.} & \multicolumn{1}{l}{All convs.}\\
\midrule
\midrule
Breakdowns & 33.4 & 43.1 & 37.1 \\
Design inadequacies & 11.1 & 19.7 & 14.4 \\
Employee service & 18.0 & 4.9 & 13.0 \\
Delays \& cancellations & 13.9 & 9.5 & 12.2 \\
Product information & 11.2 & 12.7 & 11.8 \\
Other & 9.0 & 9.2 & 9.0 \\
Envir. \& consum. health & 2.4 & 0.9 & 1.8 \\
Product quality & 1.1 & 0.1 & 0.7 \\
\bottomrule
\end{tabular}
\end{table}

\begin{figure}[ht!]
    \makebox[\textwidth][c]{\includegraphics[width=0.8\linewidth]{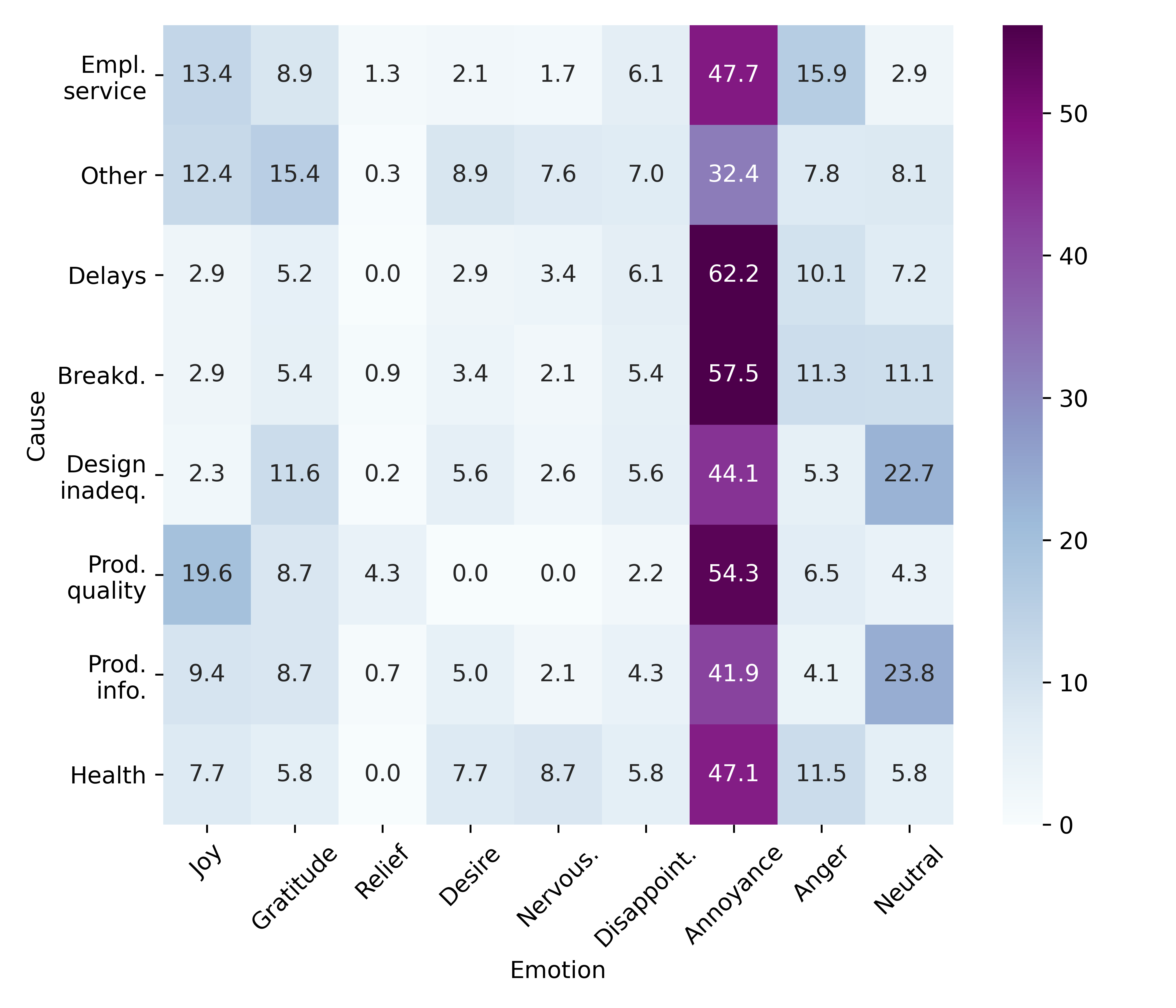}}
    \caption{Heatmap showing the distribution (in~\%) of causes across the different emotion clusters.}
    \label{fig:emo-cause}
\end{figure}

Figure~\ref{fig:emo-cause} explores how causes are related to emotion clusters. As 99.1\% of all causes are mentioned in the first customer turn, we extracted the emotions from that first customer turn for all 3,539 conversations that are labelled as subjective and contain a cause. The heatmap in Fig.~\ref{fig:emo-cause} is further normalized over causes, so that its rows can be compared with one another. Given a cause, the figure shows the frequencies (in~\%) with which this cause co-occurs with the different emotions in the first customer turn. This way, we see that the emotion \textit{annoyance} is the most frequent emotion across all causes. However, we can also notice some subtle differences between the cause categories. For example, the causes \textit{other} and \textit{product quality} are often linked to positive emotions such as \textit{joy} and \textit{gratitude}. A similar argument can be made for \textit{employee service}, although this cause is in 15.9\% of the cases also associated with the emotion \textit{anger} (a more extreme negative emotion than \textit{annoyance}). Finally, the causes \textit{design inadequacies} and \textit{product information} are in their turn often associated with a \textit{neutral} sentiment.


\subsubsection{The role of response strategies in emotion trajectories}\label{resp-emo}

\begin{figure}[h!]
    \makebox[\textwidth][c]{\includegraphics[width=0.8\textwidth]{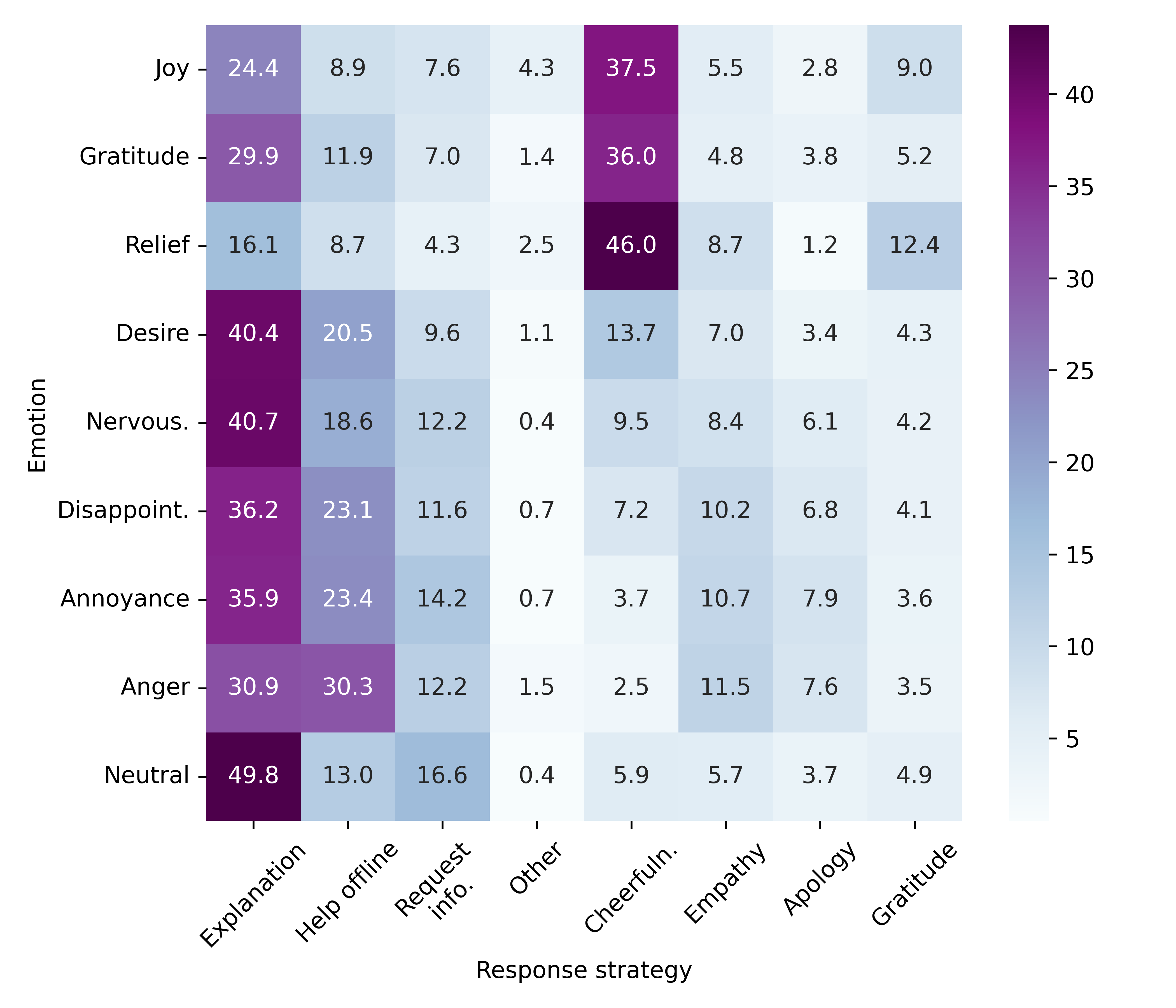}}
    \caption{Heatmap showing the distribution of response strategies in answer to a customer turn with a given emotion, for each of the emotion clusters.}
    \label{fig:emo-resp}
\end{figure}

In this part, we investigate the influence of response strategies on the emotion trajectory during the conversation. Remember that with response strategies, we refer to the emotional and informative techniques customer service agents apply in their responses. We first provide an overview of the frequencies of the different response strategy labels in Table~\ref{tab:freq-resp}. The frequencies in this table are calculated on turn level, but frequencies on tweet level can be found in Table~\ref{tab:freq-resp-tweet} in the Appendix. We also relate the response strategies to the customer emotions in Figure~\ref{fig:emo-resp}. As we only annotated response strategies in case a dialogue is subjective, the data in Table~\ref{tab:freq-resp} and Figure~\ref{fig:emo-resp} are calculated on the 5,299 conversations that meet this condition. 

\begin{table}[h!]
\caption{Frequencies (in \%) of response strategies on turn level.}
\label{tab:freq-resp}
\centering
\begin{tabular}{lr}
\toprule
Response strategy & \multicolumn{1}{l}{Turns}\\
\midrule
\midrule
Explanation & 36.9 \\
Help offline & 19.0 \\
Request information & 12.9 \\
Cheerfulness & 10.9 \\
Empathy & 8.2 \\
Apology & 5.7 \\
Gratitude & 4.4 \\
Other & 2.0 \\
\bottomrule
\end{tabular}
\end{table}

Table~\ref{tab:freq-resp} shows that informative response strategies are clearly more present than emotional ones. This result is not that surprising considering the goal-oriented nature of customer service dialogues in which clients contact companies with a particular purpose/issue to be solved. The response strategy \textit{explanation} is by far the most popular label, since it is used in more than a third of the operator turns. Moreover, in almost a fifth of the operator turns, employees offer to deliver \textit{help offline}. This trend is confirmed by other researchers who observe that on public social media channels, most companies often attempt to redirect their customers to private channels for the purpose of complaint handling to prevent negative word-of-mouth~\citep{Einwiller2015,VanHerck2020}. The third most frequent informative response strategy is \textit{request information} which occurs in 12.9\% of the operator turns. As for emotional response strategies, the most frequent category is \textit{cheerfulness} (10.9\%), followed by \textit{empathy} (8.2\%) and \textit{apology} (5.7\%). The fourth emotional response strategy \textit{gratitude} only manifests itself in 4.4\% of the operator turns.

In Section~\ref{emo-traj-evo}, we noticed that operators generally succeed at either retaining or transferring customer emotions in/to positive \textit{valence} values. To explore which response strategies operators apply to achieve this goal, we plotted the heatmap in Figure~\ref{fig:emo-resp}. Given an emotion cluster in a customer turn, the figure shows the frequencies (in~\%) of the different response strategies the operator applies in the next turn for that emotion. The frequencies in the heatmap are further normalized over the different emotions to allow for comparisons between rows. From this figure, we see a clear division between emotions that are associated with unresolved customer intents such as questions or ongoing problems (ranging the row of \textit{desire} to the row of \textit{anger}, and including the \textit{neutral} label for questions) and emotions linked to the fulfillment of these intents (namely~\textit{relief}, \textit{gratitude}, and \textit{joy}). As in Table~\ref{tab:freq-resp}, the \textit{explanation} category is frequent across all emotion classes, but it is clearly more present when the previous emotion is linked to unfulfilled intents than fulfilled intents. Other emotions associated with unfulfilled intents are \textit{help offline} and \textit{request information}, with \textit{help offline} being especially utilized after a customer expresses \textit{anger}, \textit{annoyance} or \textit{disappointment}. As for the emotions associated with fulfilled intents, the response strategy~\textit{cheerfulness} occurs most frequently after these emotions, followed by \textit{explanation} and, to a lesser extent, \textit{help offline} and \textit{gratitude}.

\section{Discussion} \label{sec5}
Before concluding this paper, we critically reflect upon our contributions. The core advantages of our novel dataset along with its limitations are described in Section~\ref{disc1}. We also emphasize the future application potential of the EmoTwiCS dataset in both research on NLP and beyond. To this end, Section~\ref{disc2} provides some suggestions on the different types of open research questions and predictive modelling tasks that can be investigated with EmoTwiCS.

\subsection{Advantages and limitations of the EmoTwiCS corpus}\label{disc1}

In this section, we outline our main contributions along five keywords: descriptors, target language, domain, communication channel, and analytical insights. The same keywords can, however, also be used to discuss limitations of our current research and define avenues for future research.

\textsc{Descriptors} — We designed an extensive annotation framework~\citep{Labat2020} that was applied to the EmoTwiCS corpus. Customer emotions are annotated in a fine-grained fashion along a categorical and dimensional framework. Given the extensive emotion taxonomy (28 categories), its clustering potential (see Section~\ref{iaa-clust}), and dimensional \textit{valence}-\textit{arousal}-\textit{dominance} scores, EmoTwiCS can more easily be compared (i) across different domains and (ii) with datasets containing less elaborate annotations. At the same time, we noticed that raising the granularity of our annotation approach translated into lower IAA results. Although we managed to partially resolve this issue by clustering emotion labels, the problem seems to be re-occurring in other research on emotion detection, underscoring the difficult and subjective nature of the task as part of ongoing research. Besides customer emotions, we added two extra annotation layers to our framework, namely causes and response strategies. The inclusion of both layers leads to some interesting application potential in different fields (see Section~\ref{disc2} for some examples).

\textsc{Target language} — ERC is mostly performed on English datasets, while many other prominent languages are currently not yet represented. For Dutch ERC, there is only one publicly released dataset available called deLearyous~\citep{Vaassen2012} which contains 11 conversations that are all grounded in the same event. The introduction of EmoTwiCS fills the existing research gap and aids the NLP community working on Dutch. We are, however, aware that this resource might be somewhat less interesting to the international community and that, at the same time, many work remains to be done for other low-resource languages concerning ERC. In future research, we therefore plan to extend the EmoTwiCS database to English and other low-resource languages in terms of conversational data for emotion analysis, so that the resulting corpora can be used in crosslingual and multilingual machine learning experiments.

\textsc{Domain} — We present EmoTwiCS as the first publicly available dataset for the task of ERC in the closed-domain of customer service. Even though we restrict ourselves to a closed-domain, the domain itself is highly prominent in a number of industrial applications. Moreover, we ensured EmoTwiCS's cross-domain transferability potential by annotating emotions along large fine-grained categorical and dimensional frameworks (see supra).

\textsc{Communication channel} — The EmoTwiCS data is scraped from Twitter, an upcoming medium to deliver customer service which is becoming increasingly important to businesses. From a practical point of view, Twitter corpora can more easily be made publicly available to the research community than the private conversational databases of companies. Even though both types of data contain customer service conversations, there exist some semantic and formal differences between the two. 
To illustrate the semantic differences, we compare our analysis with \citet{Guibon2021} who studied emotions in a live chat customer service. We notice that EmoTwiCS not only has more expressions of emotions, but when comparing the ratio of negatively to positively connotated emotions assigned to utterances, we find that EmoTwiCS contains over two times more negative emotions than the live chat dataset.\footnote{To calculate these ratios for EmoTwiCS, we relied on Table~\ref{tab:ext-freq-emo}. In EmoTwiCS, the ratio of negatively to positively connotated emotions equals $53/22.6$ or 2.35, while in \citet{Guibon2021}'s dataset, this ratio is $4.907/4.966$ or 0.99 (see Figure 1 in~\citet{Guibon2021}).} A formal difference between the two types of resources is the difference in length of the conversations. Customer service conversations on Twitter are generally shorter, making it harder to study transitions in the emotion trajectories they express. Their shorter length is probably a result of most companies attempting to mainly redirect their customers to private channels for the purpose of complaint handling~\citep{Einwiller2015,VanHerck2020}. In future research, we will therefore collaborate with companies that engage in customer service interactions and apply our existing annotation framework to their conversational datasets. We also plan to collect an `artificial' dataset with longer client-operator interactions through Wizard of Oz (WOZ) experiments.

\textsc{Analytical insights} — We performed an in-depth analysis on the data in EmoTwiCS. Not only did we investigate emotions in isolated messages, but we also studied them as dynamic attributes as part of a trajectory (along with causes and operator response strategies). Our analysis is not only informative for the development of novel machine learning systems, but it is also useful to other research disciplines such as discourse analysis, communication science, marketing research. Moreover, a similar principle applies to companies that engage with their customers, as this analysis has implications in terms of service delivery, marketing, CRM, and human resources. Nevertheless, given the differences in scope of the previously mentioned fields, more analyses can be conducted in the future. Section~\ref{disc2} gives therefore some brief suggestions on open research questions that can be addressed with the EmoTwiCS corpus. 

\subsection{Outlook on open research questions and predictive modelling with EmoTwiCS}\label{disc2}

We now turn our attention to some concrete suggestions on the different types of predictive modelling tasks and open research questions that can be addressed with EmoTwiCS. The most coarse-grained prediction tasks with direct use in customer service applications involve recognizing subjectivity and distinguishing between positive emotions (e.g., for detecting net promoters) and negative emotions (prediction of churn, detractors, and human handover). The EmoTwiCS annotation scheme further allows to not only model more fine-grained emotions, but also operator response strategies and events triggering emotions.

Fine-grained emotion detection on the level of tweets or utterances, given the preceding context of the conversation, is expected to be more difficult than predicting mere sentiment. Nevertheless, it is one of the key tasks that EmoTwiCS was designed for. We consider the joint prediction of emotions and their causes (not always present) to be an important task as well, especially since this is one of the more recent developments in the field of ERC.
In light of the analysis in Section~\ref{emo-traj-evo}, a more advanced and even sparser, yet highly interesting task is the prediction of emotion evolution throughout conversations. This may be cast, for example, as predicting whether the valence will increase or the arousal will decrease from one turn to the next, or from the first to the final customer turn. Alternatively, particular changes in customer emotions (such as those shown in Figure~\ref{fig:emo-turn-to-turn}) could be modelled explicitly. Finally, predicting the next customer emotion given the conversation history up to the current operator's response also makes for a challenging yet practically useful prediction task.

Besides these predictive modelling tasks, EmoTwiCS can also be applied to other domains to tackle unresolved research questions. For example, the dataset can be used in the field of digital business communication to tap into an ongoing research trend and investigate cross-message patterns~\citep{Grewal2021} that go beyond the individual message. EmoTwiCS might also be an interesting resource to the discourse analysis community to study the pragmatic features of customer service interactions on Twitter~\citep[see][]{VanHerck2022}. Moreover, our dataset might be of value to the domains of marketing and management research to gain insights on how customer service interactions can be optimized and how, in the long run, such insights will improve service operations (e.g., service delivery~\citet{Rafaeli2017}, human resource management~\citet{Hamilton2020}). 

\section{Conclusion} \label{sec6}
This paper presents the EmoTwiCS database as a novel resource to model emotion trajectories in 9,489 Dutch customer service dialogues on social media. With emotion trajectories we refer to the (i) emotions of customers (annotated along large categorical and dimensional frameworks), (ii) the events triggering these emotions, and (iii) operator response strategies. In our IAA study, we achieve results that are substantial and comparable to related research, indicating the high quality of our corpus. EmoTwiCS is the first publicly released corpus in its kind and taps in on various research gaps and ongoing research trends. Research gaps are addressed by focusing on Dutch as a low-resource language for ERC and on customer service as a prominent business-related application domain. An in-depth analysis is performed on (i) the distribution of customer emotions in isolated tweets and on (ii) the evolution of dynamic emotion trajectories in a conversational setting. The results of this analysis contribute not only to ongoing interdisciplinary research on customer service, but have also practical implications to business management in terms of service delivery, marketing, CRM, and human resources. Finally, EmoTwiCS has the potential to play an important role in future NLP research such as ERC (see Section~\ref{disc2}). We hypothesize that advances in terms of predictive modelling of emotional content in customer service conversations will in the long-term result in an increase in quality of human-human and human-machine customer service interactions. 

\begin{acknowledgements}
This research received funding from the Flemish Government under the Research Program Artificial Intelligence - 
174U01222 (2022). The research also received funding from the Research Foundation Flanders (FWO-Vlaanderen) with grant number 1S96322N. We would also like to thank the anonymous reviewers for their constructive and insightful feedback.
\end{acknowledgements}

\bibliographystyle{spbasic}
\bibliography{bibliography}

\clearpage\section*{Appendix}

\begin{table}[h!]
\caption{Average Fleiss' $\kappa$ for the individual emotion labels (see Table~\ref{tab:iaa-emotions}) including/excluding \textit{neutral}. The average is in both cases weighted over annotations and tweets, respectively.}
\label{tab:iaa-emotions-avg}
\centering
\begin{tabular}{llr}
\toprule 
Labels & Weighting method & Fleiss' $\kappa$ \\
\midrule
\midrule
Emotions incl. \textit{neutral} & Weighted over annots. & 0.523 \\
Emotions incl. \textit{neutral} & Weighted over tweets & 0.458 \\
Emotions excl. \textit{neutral} & Weighted over annots. & 0.421 \\
Emotions excl. \textit{neutral} & Weighted over tweets & 0.370 \\
\bottomrule
\end{tabular}
\end{table}

\begin{table}[h!]
\caption{Average Fleiss' $\kappa$ for emotion clusters (see Table~\ref{tab:emo-clusters}) including/excluding the \textit{neutral} cluster. The average is in both cases weighted over annotations and tweets, respectively.}
\label{tab:iaa-clusters-avg}
\centering
\begin{tabular}{llr}
\toprule 
Labels & Weighting method & Fleiss' $\kappa$ \\
\midrule
\midrule
Emotion clusters incl. \textit{neutral} & Weighted over annots. & 0.649\\
Emotion clusters incl. \textit{neutral} & Weighted over tweets & 0.623 \\
Emotion clusters excl. \textit{neutral} & Weighted over annots. & 0.585 \\
Emotion clusters excl. \textit{neutral} & Weighted over tweets & 0.558 \\
\bottomrule
\end{tabular}
\end{table}

\begin{table}[h!]
\caption{Frequency (in \%) of VAD scores on tweet and turn level. Frequencies are calculated on the subset of subjective conversations.}
\label{tab:ext-freq-vad}
\centering
\begin{tabular}{ll|rr}
\toprule
& & \multicolumn{1}{l}{Tweets} & \multicolumn{1}{l}{Turns} \\ 
\midrule
\midrule
\multirow{5}{*}{Valence} & 1 & 29.1 & 28.9 \\ 
& 2 & 28.3 & 28.9 \\ 
& 3 & 26.8 & 26.0 \\ 
& 4 & 11.0 & 11.3 \\ 
& 5 & 4.8 & 4.9 \\ 
\midrule
\multirow{5}{*}{Arousal} & 1 & 0.1 & 0.1 \\ 
& 2 & 3.2 & 3.2 \\ 
& 3 & 39.8 & 39.2 \\ 
& 4 & 50.4 & 50.9 \\ 
& 5 & 6.5 & 6.5 \\ 
\midrule
\multirow{5}{*}{Dominance} & 1 & 6.2 & 6.3 \\ 
& 2 & 19.8 & 19.9 \\ 
& 3 & 43.7 & 43.4 \\ 
& 4 & 28.4 & 28.5 \\ 
& 5 & 1.9 & 1.9 \\ 
\bottomrule
\end{tabular}
\end{table}

\begin{table}[h!]
\caption{Frequency (in \%) of customer emotions on tweet and turn level. Frequencies are calculated on the subset of subjective conversations.}
\label{tab:ext-freq-emo}
\centering
\begin{tabular}{lrrlrr} 
\toprule
Emotion cluster & \multicolumn{1}{l}{Tweets} & \multicolumn{1}{l}{Turns} &
Emotion label &  \multicolumn{1}{l}{Tweets} &
\multicolumn{1}{l}{Turns} \\
\midrule
\midrule
Annoyance & 37.9 & 36.9 & 
Annoyance & 26.0 & 26.0 \\ 
          & & & 
          Disapproval & 12.1 & 12.3 \\ 
\midrule
Neutral & 20.3 & 15.3 & 
Neutral & 15.4 & 14.3 \\ 
        & & & 
        Confusion & 3.7 & 3.7 \\ 
        & & & 
        Curiosity & 0.9 & 0.9 \\ 
        & & & 
        Realization & 0.5 & 0.5 \\ 
        & & & 
        Remorse & 0.3 & 0.3 \\ 
        & & & 
        Surprise & 0.3 & 0.3 \\ 
        & & & 
        Embarrassment & 0.1 & 0.1 \\ 
        & & & 
        Pride & 0.1 & 0.1 \\ 
\midrule
Gratitude & 13.5 & 15.4 & 
Gratitude & 12.8 & 13.1 \\ 
\midrule
Joy & 7.8 & 8.6 & 
Approval & 2.0 & 2.1 \\ 
    & & & 
    Joy & 1.9 & 2.0 \\ 
    & & & 
    Amusement & 1.8 & 1.8 \\ 
    & & & 
    Admiration & 1.4 & 1.5 \\ 
    & & & 
    Excitement & 0.7 & 0.7 \\ 
    & & & 
    Love & 0.4 & 0.4 \\ 
\midrule
Anger & 7.8 & 8.4 & 
Anger & 5.7 & 5.8 \\ 
      & & & 
      Disgust & 1.9 & 1.9 \\ 
\midrule
Disappointment & 5.3 & 6.3 & 
Disappointment & 3.6 & 3.6 \\ 
               & & & 
               Sadness & 1.5 & 1.5 \\ 
\midrule
Desire & 4.1 & 5.2 & 
Desire & 3.2 & 3.2 \\ 
       & & & 
       Optimism & 0.8 & 0.8 \\ 
\midrule
Nervousness & 2.0 & 2.2 & 
Nervousness & 1.4 & 1.4 \\ 
            & & & 
            Fear & 0.5 & 0.5 \\ 
\midrule
Relief & 1.3 & 1.7 & 
Caring & 0.8 & 0.8 \\ 
       & & & 
       Relief & 0.5 & 0.5 \\ 
\bottomrule
\end{tabular}
\end{table}

\begin{table}[h!]
\caption{Frequencies (in \%) of response strategies on tweet level.}
\label{tab:freq-resp-tweet}
\centering
\begin{tabular}{lr}
\toprule
Response strategy &  \multicolumn{1}{l}{Tweets}\\
\midrule
\midrule
Explanation & 39.1\\
Help offline & 18.3\\
Request information & 12.6\\
Cheerfulness & 10.5\\
Empathy & 7.9\\
Apology & 5.5\\
Gratitude & 4.3\\
Other & 1.9\\
\bottomrule
\end{tabular}
\end{table}
%
%
%
%
%
%
%
\end{document}